\documentclass[twoside,leqno,twocolumn]{article}
\UseRawInputEncoding 
\usepackage[letterpaper]{geometry}
\usepackage{ltexpprt}
\usepackage{algorithm}
\usepackage[noend]{algorithmic}
\usepackage{tikz}
\usepackage{multirow}

\usepackage{marvosym}
\usepackage{cite}
\usepackage{amsmath,amssymb,amsfonts}
\usepackage{graphicx}
\usepackage{textcomp}
\usepackage{xcolor}
\usepackage{makecell}
\usepackage{subcaption}
\usepackage[showframe=true]{}
\usepackage{amsmath}
\usepackage{hyperref}
\usepackage{booktabs}
\usepackage{enumitem}
\usepackage{breqn}
\usepackage{amsfonts,amssymb}
\usepackage{pdfpages}
\newtheorem{definition}{Definition}

\begin{document}

\newcommand\relatedversion{}

\title{\Large Hierarchical Superpixel Segmentation via Structural Information Theory\relatedversion}

\author{
Minhui Xie\thanks{School of Cyber Science and Technology, Beihang University, China. \{xieminhui, penghao, zengguangjie\}@buaa.edu.cn, lp2024626@gmail.com, lp\_academy@163.com.} 
\and Hao Peng$^*$\textsuperscript{\Letter}
\and Pu Li$^*$ 
\and Guangjie Zeng$^*$ 
\and Shuhai Wang\thanks{College of Information Science and Technology, Shijiazhuang Tiedao University, China. wsh36302@126.com.}
\and Jia Wu\thanks{Macquarie University, Australia. jia.wu@mq.edu.au.}
\and Peng Li$^*$ 
\and Philip S. Yu\thanks{University of Illinois at Chicago, Illinois, USA. psyu@uic.edu.}
}

\date{}

\maketitle

% Copyright Statement
% When submitting your final paper to SIAM proceedings, please include
% the appropriate copyright in the footer of the paper.  The copyright added should be
% consistent with the copyright selected on the copyright form submitted with the paper.
% Please note that "20XX" should be changed to the year of the meeting.

% Default Copyright Statement
\fancyfoot[R]{\scriptsize{Copyright \textcopyright\ 2025 by SIAM\\
Unauthorized reproduction of this article is prohibited}}

% Depending on which copyright you agree to when you sign the copyright form, the copyright
% can be changed to one of the following after commenting out the default copyright statement
% above.

%\fancyfoot[R]{\scriptsize{Copyright \textcopyright\ 20XX\\
%Copyright for this paper is retained by authors}}

%\fancyfoot[R]{\scriptsize{Copyright \textcopyright\ 20XX\\
%Copyright retained by principal author's organization}}

%\pagenumbering{arabic}
%\setcounter{page}{1}%Leave this line commented out.
%Its answer can augment the event data for representation learning

\begin{abstract} \small\baselineskip=9pt
Superpixel segmentation is a foundation for many higher-level computer vision tasks, such as image segmentation, object recognition, and scene understanding.
Existing graph-based superpixel segmentation methods typically concentrate on the relationships between a given pixel and its directly adjacent pixels while overlooking the influence of non-adjacent pixels. 
These approaches do not fully leverage the global information in the graph, leading to suboptimal segmentation quality.
To address this limitation, we present SIT-HSS, a hierarchical superpixel segmentation method based on structural information theory.
Specifically, we first design a novel graph construction strategy that incrementally explores the pixel neighborhood to add edges based on 1-dimensional structural entropy (1D SE).
This strategy maximizes the retention of graph information while avoiding an overly complex graph structure.
Then, we design a new 2D SE-guided hierarchical graph partitioning method, which iteratively merges pixel clusters layer by layer to reduce the graph's 2D SE until a predefined segmentation scale is achieved.
Experimental results on three benchmark datasets demonstrate that the SIT-HSS performs better than state-of-the-art unsupervised superpixel segmentation algorithms.
The source code is available at \url{https://github.com/SELGroup/SIT-HSS}.

\let\thefootnote\relax\footnotetext[1]{\textsuperscript{\Letter} Corresponding author.}

\end{abstract}

\section{Introduction}
Superpixel segmentation is a key technique in computer vision, intending to group pixels into compact regions with a consistent semantic context~\cite{STUTZ20181}.
It usually works as the basis for higher-level computer vision tasks by reducing the number of primitives~\cite{Ren2003LearningAC}.
Various downstream applications can benefit from superpixel segmentation, including image segmentation~\cite{lei2018superpixel,ng2023fuzzy}, image classification~\cite{sellars2020superpixel,zhao2023hyperspectral}, saliency detection~\cite{zhang2019spatiotemporal,qiu2024superpixel}, and 3D reconstruction~\cite{huang2016robust,zhang2024supernerf}.

\begin{figure}[t]
    \center
    \includegraphics[width=0.48\textwidth]{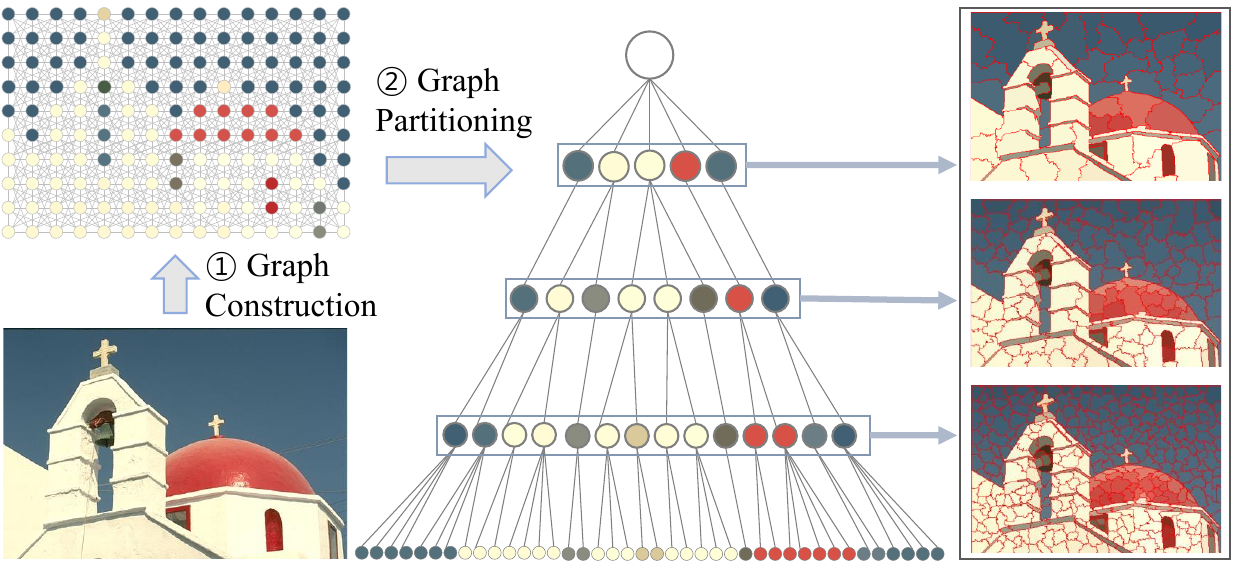}
    \caption{ An illustration of superpixel segmentation using SIT-HSS.
It performs superpixel segmentation by first constructing a weighted graph, then obtaining an optimal encoding tree, and finally achieving multi-scale superpixel segmentation.}
 \label{fig_toy}
 \vspace{-3mm}
\end{figure}

The requirements of superpixel segmentation vary for different downstream application scenarios, where some commonly agreed requirements include \textbf{Connectivity}, \textbf{Boundary Adherence}, \textbf{Compactness}, and \textbf{Computational Efficiency}.
In the last two decades, superpixel segmentation has made considerable progress with the emergence of numerous techniques, such as watershed-based \cite{yuan2020watershed}, density-based \cite{vedaldi2008quick}, path-based \cite{tang2012topology}, clustering-based \cite{SLIC}, graph-based \cite{SH} and more recently, deep learning-based methods \cite{FCN}.
Among them, density-based methods cannot determine the number of resulting superpixels, whereas path-based methods generate superpixels with poor compactness \cite{STUTZ20181}.
Although some methods \cite{chang2024hierarchical} focus on a single requirement, most involve trade-offs between different requirements to cover a wide range of downstream applications.
Graph-based methods construct graphs from the given image and partition these graphs into subgraphs as superpixels.
This graph construction step helps them control the trade-off between boundary adherence and compactness.

Various graph-based methods have been proposed since NCut \cite{shi2000normalized} is utilized as a superpixel segmentation technique for image classification \cite{Ren2003LearningAC}.
They typically map the image as a graph and optimize a specific objective, such as the entropy rate \cite{ERS} or normalized cuts \cite{Ren2003LearningAC} to achieve graph partitioning for superpixel segmentation.
Recently, the hierarchy of superpixels has been addressed, as the multi-resolution representation of hierarchical superpixel segmentation is believed to benefit downstream applications \cite{SH, CRTrees}.
Despite the significant advancement achieved by these methods, existing graph-based methods overlook the relationships of non-adjacent pixels.
In the graph construction step, they connect only pixels to their directly adjacent pixels.
These adjacent pixel graphs do not fully capture the relationships between pixels and inevitably impede overall segmentation performance.

To address the aforementioned issues, we present SIT-HSS, a hierarchical superpixel segmentation method based on structural information theory.
As illustrated in Fig.~\ref{fig_toy}, we achieve superpixel segmentation by graph partitioning on a pixel graph, where the graph is constructed by maximizing 1D SE, and graph partitioning is conducted by minimizing 2D SE.
Maximizing 1D SE retains more information between pixels of the image in the graph, delivering high-quality superpixel segmentation with accurate capture of object boundaries.
Firstly, we construct an undirected pixel graph from the image. 
In this graph, graph nodes represent pixels, and weighted edges connect spatially close pixels, measuring their colour and spatial space similarities.
To extract adequate information from the image, we devise a novel graph construction strategy by maximizing the 1D SE of the graph.
This strategy calculates the graph's 1D SEs with increasing search radii and selects the radius where the 1D SE plateaus.
Pixels connect to neighbour pixels within this radius, extracting sufficient information while avoiding complex, dense graphs.
Secondly, we propose a hierarchical graph partitioning algorithm based on merging using 2D SE minimization.
This algorithm iteratively merges the partition pair with the most significant decrease in 2D SE until the number of partitions matches the pre-defined superpixel segmentation scale.
Additionally, the merging is performed only on pairs of partitions directly adjacent to the image, ensuring the connectivity of the superpixels generated.

We comprehensively evaluated the SIT-HSS in three datasets compared to nine baselines.
Experimental results demonstrate that SIT-HSS outperforms state-of-the-art superpixel segmentation methods regarding segmentation accuracy while maintaining competitive efficiency.
The contributions of this paper are summarized below.

$\bullet$
An interpretable hierarchical superpixel segmentation method based on structural information theory, namely SIT-HSS, is presented.

$\bullet$
A novel graph construction strategy based on 1D SE maximization is devised to retain adequate information from the image while avoiding complex, dense graphs.

$\bullet$
A hierarchical graph partitioning algorithm based on 2D SE minimization is proposed to achieve efficient and accurate superpixel segmentation.

$\bullet$
Extensive comparative experiments and visualizations are conducted on three representative datasets, demonstrating the state-of-the-art (SOTA) performance of SIT-HSS on superpixel segmentation.
\section{Preliminaries}\label{sec: Preliminary}

\begin{figure*}[t]
    \center
    \includegraphics[width=1\textwidth]{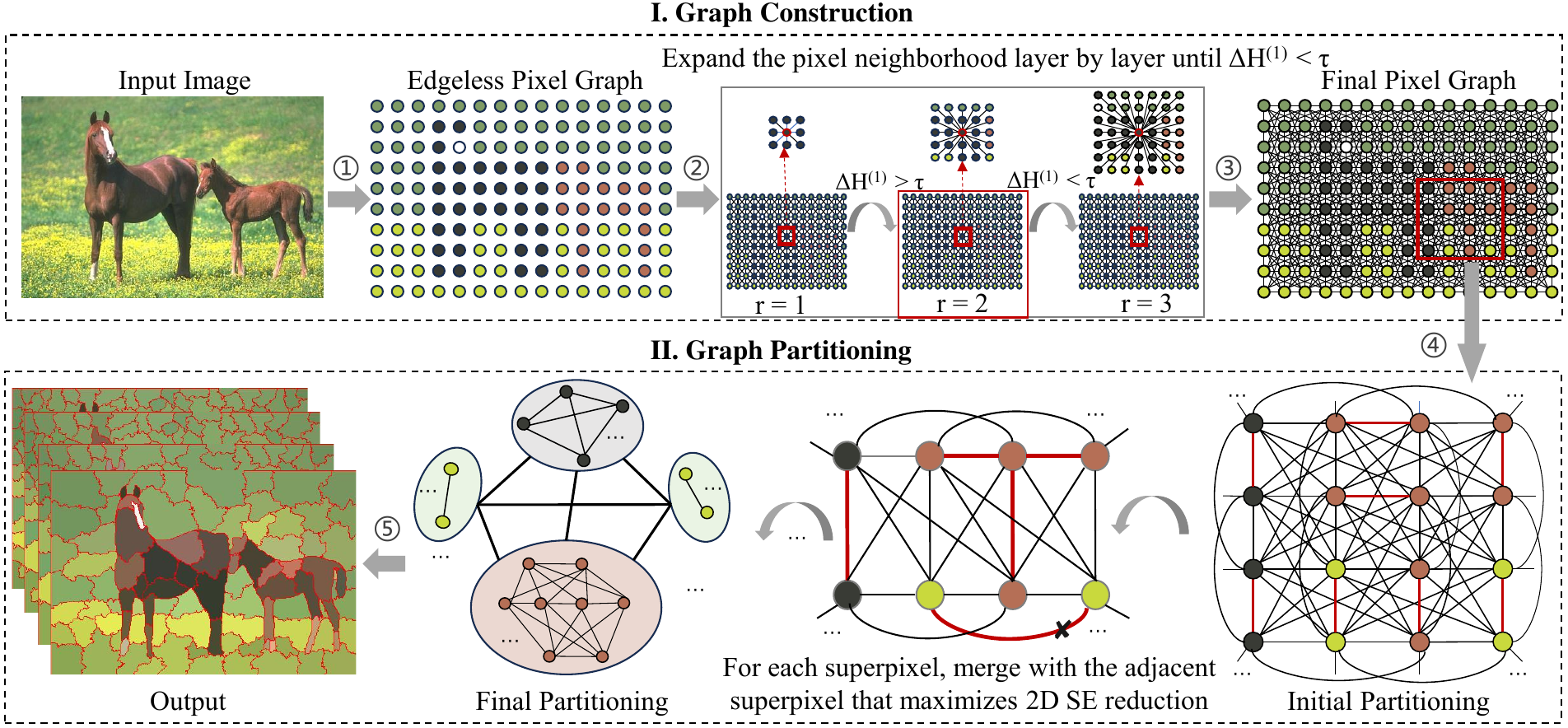}
	\vspace{-1em}
	\caption{The framework of the SIT-HSS. I and II represent the weighted graph construction and partitioning processes, respectively. We add edges to the edgeless pixel graph by progressively expanding the neighbourhood. Then, for each superpixel, we iteratively merge neighbouring superpixel pairs that maximally reduce the 2D SE until the target size is reached.}
	\label{fig_framework}
    \vspace{-4mm}
\end{figure*}

In this section, we briefly introduce the structural information theory on which our SIT-HSS is based.
First proposed by Li and Pan~\cite{Li2016}, structural information theory comprehensively defines structural entropy, quantifying graph complexity and connectivity patterns through encoding trees.
It has achieved successful applications in many fields, including skin lesion segmentation \cite{zeng2023unsupervised}, graph clustering \cite{zeng2024scalable, sunlsenet}, classification \cite{yang2024hierarchical} and graph structure learning \cite{zou2023se}, social bot detection \cite{pengunsupervised2024}, social event detection \cite{cao2023hierarchical}, and reinforcement learning \cite{zeng2023effective, zeng2024effective}.
The structural entropy of a graph is defined as the minimum average number of communities required to encode the graph nodes reachable in a random walk.
The detailed definitions of encoding trees and structural entropy are given below.

\begin{definition}
(Encoding Tree). 
Given a graph $G = (V, E,\mathbf{W})$, an encoding tree \(T\) is a hierarchical partition of the graph \(G\). 
The encoding tree satisfies the following conditions:

1)The root node \(\lambda\) of the encoding tree corresponds to the entire set of graph nodes, \(T_{\lambda} = V\). 
Each node \(\alpha\) in the tree is associated with a subset of graph nodes \(T_{\alpha} \subseteq V\). 

2)For any leaf node \(\gamma\), \(T_{\gamma}\) contains exactly one graph node, i.e., \(T_{\gamma} = \{v\}\), where \(v \in V\). 
For any node \(\alpha\) in the tree, its children are denoted as \(\beta_1, ..., \beta_k\), so \((T_{\beta_1}, ..., T_{\beta_k})\) forms a partitioning of \(T_{\alpha}\).

3)Each tree node \(\alpha\) has a height denoted as \(h(\alpha)\). For any leaf node \(\gamma\), \(h(\gamma) = 0\); 
for a node \(\alpha^{-}\) (the parent of \(\alpha\)), \(h(\alpha^{-}) = h(\alpha) + 1\). 
The height of the tree \(T\), denoted as \(h(T)\), is defined as the maximum height of all nodes, i.e., \(h(T) = \max_{\alpha \in T} \{h(\alpha)\}\).
\end{definition}

\begin{definition}
(Structural Entropy). The structural entropy of a graph \(G\) with respect to an encoding tree \(T\) is defined as:
\begin{equation}
    H^T(G) = -\sum_{\alpha \in T, \alpha \neq \lambda} \frac{g_{\alpha}}{\mathcal{V}_G} \log \frac{\mathcal{V}_\alpha}{\mathcal{V}_{\alpha^{-}}},
\end{equation}
where \(g_{\alpha}\) is the cut, i.e., the weighted sum of edges with exactly one endpoint in \(T_{\alpha}\). \(\mathcal{V}_\alpha\), \(\mathcal{V}_{\alpha^{-}}\), and \(\mathcal{V}_G\) denote the volumes, i.e., the sums of node degrees within \(T_{\alpha}\), \(T_{\alpha^{-}}\), and \(G\), respectively.

The structural entropy of different dimensions corresponds to encoding trees of different heights, reflecting various levels of structural information in the graph. 
The \(D\)-dimensional SE \(H^{(D)}(G)\) of \(G\) is defined as:
\begin{equation} 
H^{(D)}(G) = \min_{\forall T: h(T) = D} \{H^T(G)\},
\end{equation}
which is realized by obtaining an optimal encoding tree of height \(D\) that minimizes the interference caused by noise or random variations. 
\end{definition}
\section{Methodology}\label{sec: Methodology}
Superpixel segmentation groups pixels with similar colours and proximity.
We approach this as an interpretable graph partitioning problem, where pixels are represented as graph nodes, and closely related nodes are grouped into partitions as superpixels.
As illustrated in Fig.~\ref{fig_framework}, our proposed SIT-HSS framework consists of two modules: graph construction and partitioning.
In the graph construction module, we build a multilayer neighbourhood graph based on 1D SE maximization.
This entropy maximization strategy enhances the expressiveness of graphs by retaining more information.
In the graph partitioning module, we employ a heuristic node merging strategy to partition graphs efficiently using 2D SE minimization.
We introduce each module in detail in the following subsections.
\subsection{Problem Formalization.}
Given an image as the input, the task of superpixel segmentation can be accomplished by constructing and partitioning a weighted graph \( G = (V, E, \mathbf{W}) \), where \( V = \{v_1, \ldots, v_N\} \) represents the pixel points of the image, \( E \) denotes the edges, and the weight matrix \( \mathbf{W} \) measures the similarities between pixels.
For two pixel points \( v_i \) and \( v_j \), their distance is:  
\begin{equation}
    \boldsymbol{\rho}_{i,j} = \|c_i - c_j\|^2_2 \cdot \|s_i - s_j\|_2, 
    \label{eq_3.3}
\end{equation}
where \( c_i \) and \( c_j \) represents the color features of \( v_i \) and \( v_j \), respectively, and
 \( s_i \) and \( s_j \) represents the position features of \( v_i \) and \( v_j \), respectively.
\( \| \cdot \|_2 \) denotes Frobenius norm.
Then we normalize this distance to represent the weight \( \mathbf{W}_{i,j} \) between the two points:
\begin{equation}
    \mathbf{W}_{i,j} = \exp(-\frac{\boldsymbol{\rho}_{i,j}}{t \cdot \frac{1}{|\boldsymbol{\rho}|}\sum_{k=1}^{|\boldsymbol{\rho}|} \boldsymbol{\rho}_k}),
\label{eq_3.4}
\end{equation}
where \(|\boldsymbol{\rho}|\) is the total number of edges, \(\boldsymbol{\rho}_k\) is the distance between the endpoints of the \(k\)-th edge, and \( t \) is a normalization parameter.
% a parameter controlling graph sparsity.
The edge set \(E\) is initially empty and will be expanded by the graph construction process.
Partitioning \(G\) results in a set \(\{p_1, \ldots, p_n\}\), where \(\bigcap_{i \neq j} (p_i \cap p_j) = \emptyset\) and \(\bigcup_{i=1}^n p_i = V\).
Each partition \(p_i\) represents a segmented superpixel.

\subsection{Graph Construction.}\label{sec_graph_construction}
Existing graph-based superpixel segmentation methods focus only on the relationships between adjacent pixels and overlook the relationships between non-adjacent pixels, leading to inferior performance.
We propose a graph construction strategy based on 1D SE to retain as much information from the image as possible.
As shown in Fig.~\ref{fig_framework} (I), we construct an edgeless graph $G$ with pixels as nodes and add edges to it by progressively expanding the neighbourhood.
For each graph node $v_i$, we define the neighbourhood $\mathcal{N}_{v_i}^r$ as the set of pixels (graph nodes) within a distance of $r$ and connect $v_i$ to all nodes in $\mathcal{N}_{v_i}^r$.
When the search radius $r=1$, the neighbourhood $\mathcal{N}_{v_i}^r$ of $v_i$ consists of the eight directly adjacent pixels.
When $r$ is more considerable, $\mathcal{N}_{v_i}^r$ expands to include pixels at a more considerable distance.
The 1D SE of $G$ at a given search radius $r$ is given by:
\begin{equation}
    H^{(1)}_r(G) = - \sum_{i=1}^{|V|} \frac{d_i}{\mathcal{V}_{G}^r} \log \frac{d_i}{\mathcal{V}_{G}^r},
\label{eq_3.5}
\end{equation}
where \(\mathcal{V}_{G}^r \) is the volume of \(G\) under \( r \), and \(d_i \) is the sum of weights of edges connecting \( v_i \). 
We tend to construct a graph with the most information retained, corresponding to the graph with the largest 1D SE.
As the search radius $r$ increases, the 1D SE of $G$ typically increases.
However, a large value of $r$ leads to a dense complex $G$, making the subsequent optimization of graph partitioning difficult.
Moreover, with further increases in $r$, the rate of growth of the 1D SE often decreases.
Therefore, we select an appropriate value of \( r \) to construct a graph that retains adequate information while preventing an overly complex graph structure.
Formally, we select $r$ as the following:
\begin{equation}
    r = \arg \min_i \left\{ i \mid H_{i+1}^{(1)}(G) - H_i^{(1)}(G) < \tau \right\},
\label{eq_3.6}
\end{equation}
where \( \tau \) is a predefined SE threshold for 1D SE change.

\begin{algorithm}[t]
  \caption{Proposed SIT-HSS}
  \label{alg:Proposed SIT-HSS1}
  \begin{algorithmic}[1]
    \STATE \textbf{Input:} An image $I \in \mathbb{R}^{H \times W}$, target size $K$, $t$, $\tau$.
    \STATE \textbf{Output:} Superpixel blocks of images $\mathcal{P}$.
    \STATE // \textit{Graph Construction.} 
    \STATE $G\left(V\right) \gets$ Convert image $I$ to an edgeless graph;
    \FOR{$r \in 1,2,3,...$}
        \FOR{pixel $v_i$ in $V$}
            \FOR{$v_j$ in $\mathcal{N}_{v_i}^r$} %这里是否解释N_r是i在半径r内的像素的集合
                \STATE add $(v_i,v_j)$ to $E_r$, $\boldsymbol{\rho}_{i,j} \gets$ Eq.~\ref{eq_3.3};
            \ENDFOR
        \ENDFOR
        \STATE $\mathbf{W}_r \gets$ Eq.~\ref{eq_3.4};
        \STATE $H^{\left(1\right)}_r(G) \gets Eq.~\ref{eq_3.5}$;
        \IF{$H^{(1)}_{r} - H^{(1)}_{r-1}  \leq \tau$}
            \STATE $G \gets G\left(V,E_{r-1},\mathbf{W}_{r-1} \right)$;
            \STATE \textbf{Break};
        \ENDIF
    \ENDFOR
    \STATE // \textit{Graph Partitioning.}
    \STATE $\mathcal{P} = \{\{v_1\},\{v_2\},...,\{v_n\}\}$;
    \WHILE{True}
        \FOR{superpixel $p_i$ in $\mathcal{P}$}
            \STATE $p_{tgt}^i \gets \varnothing$, $\Delta_{max} \gets 0$;
            \FOR{adjacent superpixel $p_j$ in $\mathcal{N}_{p_i}$}
                \STATE$\Delta H^{(2)}_{p_i,p_j} \gets$ Eq.~\ref{eq_3.8};
                \IF{$\Delta H^{(2)}_{p_i,p_j} > \Delta_{max}$}
                    \STATE $p_{tgt}^i \gets p_j$, $\Delta_{max} \gets \Delta H^{(2)}_{p_i,p_j}$;
                \ENDIF
            \ENDFOR
            \STATE Merge $p_i$, $p_{tgt}^i$ to $p_i^\prime$;
            
        \ENDFOR
            \STATE Update $\mathcal{P}$ and all superpixels' volumes and cuts;
            \IF{$|\mathcal{P}| = K$}
            \STATE \textbf{Return} $\mathcal{P}$;
            \ENDIF
    \ENDWHILE
  \end{algorithmic}
\label{algorithmic}
\end{algorithm}

\subsection{Graph Partitioning.}
The 2D SE of a given graph $G$ represents the minimum SE achievable through traversing encoding trees at the height of 2, which corresponds to graph partitionings.
We perform superpixel segmentation by minimizing the 2D SE of the constructed weighted graph $G$.
As illustrated in Fig.~\ref{fig_framework} (II), we employ a method based on heuristic node merging for the 2D SE minimization.

For a weighted pixel graph $G$, the 2D SE of $G$ given by a partitioning $\mathcal{P} = \{ p_1, \ldots, p_n \} $ is defined as follow:
\begin{equation}
H^{(2)}_{\mathcal{P}}(G) = - \sum_{p_i \in \mathcal{P}} \bigg(\frac{g_{p_i}}{\mathcal{V}_G} \log \frac{\mathcal{V}_{p_i}}{\mathcal{V}_G}+ \sum_{j \in p_i} \frac{d_j}{\mathcal{V}_G} \log \frac{d_j}{\mathcal{V}_{p_i}}\bigg),
\label{eq_3.7}
\end{equation}
where \( d_j \) is the sum of edge weights connecting \( p_i \) and other graph nodes, \( \mathcal{V}_{p_i} \) and \( \mathcal{V}_G \) are the sum of node degrees in partition \( p_i \) and graph \( G \), respectively, and \( g_{p_i} \) is the sum of edge weights between nodes inside and outside partition \( p_i \).
At first, each pixel belongs to an independent partition, i.e. \( p_x = \{v_x\} \).
We minimize the 2D SE objective defined by Eq.~\ref{eq_3.7} via the merging operator~\cite{Li2016}.
For a pair of directly adjacent superpixels \( p_i \) and \( p_j \), \textit{MERGE}(\( p_i, p_j \)) adds a new superpixel \(p_i^\prime\) to \( G \), whose child nodes are the union of the child nodes of \( p_i \) and \( p_j \), and removes \( p_i \) and \( p_j \) from \( G \). 
After merging, the new partition is \( \mathcal{P}^\prime = \{ p_1, \ldots, p_i^\prime, \ldots, p_n \} \), where \( p_i^\prime = p_i \cup p_j \).
% For adjacent superpixels \(p_i\) and \(p_j\), the new partition after merging them is \( P\prime = \{ p_1, \ldots, p_i^\prime, \ldots, p_n \} \), where \( p_i^\prime = p_i \cup p_j \).
The decrease amount of \(H^{(2)}_{\mathcal{P}}(G)\) after merging is given by: 
\begin{equation}
\begin{split}
    &\Delta H^{(2)}_{p_i,p_j} = H^{(2)}_{\mathcal{P}}- H^{(2)}_{\mathcal{P}^\prime}\\
    &=\frac{1}{\mathcal{V}_G} \bigl( (\mathcal{V}_{p_i} - g_{p_i}) \log \mathcal{V}_{p_i} + (\mathcal{V}_{p_j} - g_{p_j}) \log \mathcal{V}_{p_j} \\
    &- (\mathcal{V}_{p_i^\prime} - g_{p_i^\prime}) \log \mathcal{V}_{p_i^\prime}+ (g_{p_i}+ g_{p_j}-g_{p_i^\prime}) \log {\mathcal{V}_G} \bigr),
    \label{eq_3.8}
\end{split}
\end{equation}
where \(H^{(2)}_{\mathcal{P}}\) and \(H^{(2)}_{\mathcal{P}^\prime}\) denote the 2D SE of the graph under the partitioning \(\mathcal{P}\) and \(\mathcal{P}^\prime\), respectively.
Therefore, we select a superpixel directly adjacent for each superpixel, which results in the most significant decrease in 2D SE after merging in each iteration.
This adjacent partition selection is formalized as follows:
\begin{equation}
p_{tgt}^i = \arg \max _{p_j \in \mathcal{N}_{p_i}}(\Delta H_{p_i,p_j}^{(2)}),
\label{eq_3.9}
\end{equation}
where \( p_{tgt}^i \) is the target superpixel for \( p_i \) and \( \mathcal{N}_{p_i} \) is the set of directly adjacent superpixels of \( p_i \).
At the end of each iteration, we dynamically update the volume of each partition \( \mathcal{V}_{p_i} \) and the external edge weights \( g_{p_i} \). 
Finally, the iteration process stops when the target number of superpixels, known as the target size ($K$), is reached.

\subsection{Time Complexity.}
The overall time complexity of the proposed SIT-HSS algorithm is \( O(|E|) \), as shown in Algorithm~\ref{algorithmic}, where \( |E| \) denotes the number of edges in the pixel graph. 
Specifically, the time complexity for searching the optimal radius during the graph construction phase is \( O(l_1 \cdot |V| \cdot k) = O(|E|) \) (lines 4-13), where \( l_1 \) is the number of search iterations and \( k \) is the number of neighbouring pixels. 
Additionally, the time required for graph partitioning is \( O(l_2 \cdot |\mathcal{P}| \cdot m) \leq O(|E|) \) (lines 15-26), where \(l_2\) is the number of iterations, \( |\mathcal{P}| \) (with \( |\mathcal{P}| \leq |V| \)) is the number of superpixels, and \( m \) is the average number of neighbors for each superpixel. 
Therefore, the overall time complexity of SIT-HSS is \( O(|E| + |E|) = O(|E|) \).
We further discuss the algorithm's convergence in detail in the appendix.
Notably, we employ tensor operations to optimize actual running time, allowing all superpixels to nearly parallelize the selection and merging of the optimal neighbouring pixels in each iteration.
Additionally, the process can be further accelerated through GPU processing. 
By leveraging GPUs' computing capabilities and tensor operations, SIT-HSS is competitive with the fastest existing algorithms in terms of efficiency.
\section{Experiments}\label{sec: ExperimentalSetup}
In this section, we conduct a series of extensive experiments to evaluate the performance of the SIT-HSS. 
Through these experiments, we aim to address the following four research questions:
\textbf{RQ1}: How does the proposed SIT-HSS perform on different datasets compared to the baselines? 
\textbf{RQ2}: How do the visualization results of SIT-HSS image segmentation perform, and does the algorithm exhibit interpretability?   
\textbf{RQ3}: How do the hyperparameters affect the performance of the SIT-HSS algorithm?
\textbf{RQ4}: How does SIT-HSS perform in terms of efficiency compared to baselines?

\noindent\textbf{Datasets.}
We perform experiments on three representative datasets, including the Berkeley Segmentation Dataset and Benchmark 500 (BSDS500)~\cite{BSDS50}, the Stanford Background Dataset (SBD)~\cite{SBD} and the PASCAL-S dataset~\cite{VOC}. 
The BSDS500 dataset includes 500 grayscale and colour images from natural scenes, providing an empirical foundation for image segmentation research. 
On average, each image in the BSDS500 is segmented into five distinct regions.
The SBD dataset is created by selecting complex outdoor scenes from several datasets, totalling 715 images. 
The PASCAL-S dataset contains 850 images with different object categories and sizes.

\noindent\textbf{Evaluation Metrics.}
We employ four widely used metrics~\cite{STUTZ20181} to evaluate the quality of superpixel segmentation, including Achievable Segmentation Accuracy (ASA), Boundary Recall (BR), Undersegmentation Error (UE), and Explained Variation (EV).
ASA, BR, and UE depend on expert ground truth annotations, while EV is independent of any ground truth.
Specifically, ASA measures the achievable fraction of correctly labeled superpixels, BR evaluates how well the superpixel boundaries align with the ground truth boundaries, UE assesses how much a superpixel spans multiple ground-truth segments, and EV measures the extent of variations that superpixels capture.

\noindent\textbf{Baselines.}
\textbf{1) Graph-based methods} treat each pixel as a node in a graph and generate superpixels by minimizing a cost function defined on the graph: ERS~\cite{ERS}, and DRW~\cite{DRW}.
\textbf{2) Clustering-based methods} group pixels into clusters (i.e., superpixels) and iteratively refine them until certain convergence criteria are met: SLIC~\cite{SLIC}, DBSCAN~\cite{DBSCAN}, LSC~\cite{LSC}, SNIC~\cite{SNIC}, SEEDS~\cite{SEEDS} and ETPS~\cite{ETPS}.
\textbf{3) Unsupervised deep learning-based methods} generate superpixels through end-to-end clustering without the need for labeled data: LNSNET~\cite{LNSNet}.

\noindent\textbf{Implementation Details.}
For all datasets, we set the normalization parameter $t = 0.1$.
For the BSDS500 and PASCAL-S datasets, we set the SE increment threshold $\tau = 2e-7$; for the SBD dataset, we set $\tau = 1e-6$.
All experiments are conducted on one server, which has an Intel(R) Xeon(R) Platinum 8336C CPU operating at 2.30 GHz and an RTX 4090 GPU with 24GB of memory.

\begin{table*}[t]
\renewcommand{\arraystretch}{0.99}
\centering
\caption{Quantitative results of superpixel segmentation methods when $K = 600$. The best results are \textbf{bolded}, and the second-best results are \underline{underlined}.}
\label{tab3}
\resizebox{\textwidth}{!}{%
\begin{tabular}{lcccc|cccc|cccc}
    \toprule
    \multicolumn{1}{l}{\multirow{2}{*}{Method (\%)}} 
    & \multicolumn{4}{c}{BSDS500} 
    & \multicolumn{4}{c}{SBD} 
    & \multicolumn{4}{c}{PASCAL-S}\\
    \cmidrule(r){2-5} \cmidrule(r){6-9} \cmidrule(r){10-13} 
    & ASA$\uparrow$ & BR$\uparrow$ & UE$\downarrow$ & EV$\uparrow$
    & ASA$\uparrow$ & BR$\uparrow$ & UE$\downarrow$ & EV$\uparrow$
    & ASA$\uparrow$ & BR$\uparrow$ & UE$\downarrow$ & EV$\uparrow$\\
    \midrule
    SLIC  & 95.890  & 89.903  & 3.967 & 84.456  & 94.909 & 85.721 & 5.201 & 90.863  &98.401 &87.967 &8.092  & 89.255\\ 
    SEEDS & 94.673  & \underline{97.018}  & 5.337  & 87.375 & 93.825 & \underline{97.673} &6.349  & 93.276 &98.520 &\underline{96.099}&7.668& 89.035\\ 
    ERS  & 95.699  & 94.887  &4.025 & 82.335 & 94.893 & 90.026 &5.280 & 87.005 &98.503 &95.087 &7.652  & 89.915\\ 
    ETPS & \underline{96.258}  & 91.197  & \underline{3.552}  & \underline{89.051} & \underline{95.203} & 89.002 & \underline{4.867} & \underline{94.115} &\underline{98.776} & 95.376 &\underline{7.032} & \textbf{91.052} \\ 
    DBSCAN   & 95.694 & 89.889  & 4.145  & 85.225 & 94.530 & 86.497  &5.547 & 90.827 & 98.376 & 91.668&8.090&90.042\\ 
    LSC & 95.640 & 89.745 & 4.400  & 85.823 & 94.602 & 87.141  &5.590 & 90.441 &98.524 &95.028&7.640&89.986\\ 
    SNIC & 96.122  & 89.772 & 3.820  & 86.005 & 94.900 & 85.708 &5.109 & 92.457 &98.215 & 90.449&8.278&90.543\\ 
    LNSNET & 95.705 & 90.510 & 4.197  & 82.590 & 94.506 & 88.290 &5.510  & 89.386 & 98.267 & 90.360   &8.059& 88.569\\
    % RSS & 96.443 & 93.396 &3.732  & 85.447 & 95.030 &94.432 &4.976 & 90.122 & 98.592 & 94.098  &7.588  & 89.311\\
    DRW & 95.644 & 94.577 & 4.209  & 85.110 & 94.703 & 90.118 &5.201 & 90.838 & 98.262 & 94.143  &7.809  & 89.120\\
    \midrule
    \textbf{SIT-HSS} & \textbf{96.824}  & \textbf{97.981} & \textbf{3.076}  & \textbf{89.078} & \textbf{95.250} & \textbf{99.701} & \textbf{4.729}  & \textbf{94.392} &\textbf{98.822} &\textbf{97.248}&\textbf{6.178}&\underline{90.870}\\
    relative gain & 0.566 & 0.963 & 0.476 & 0.027 & 0.047 & 2.208 & 0.138 & 0.277 & 0.046 & 1.149 & 0.854&-0.182\\
    \bottomrule
\label{tab_quantitative}
\end{tabular}%
}
\vspace{-4mm}
\end{table*}
\begin{figure*}[t]
    \centering
    \begin{subfigure}{1\linewidth}
        \centering
    \includegraphics[width=\linewidth,trim={0 1.5cm 0.4cm 1.5cm}, clip]{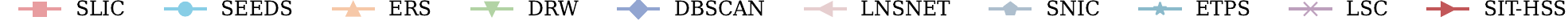}
    \end{subfigure}
    \begin{subfigure}{0.245\linewidth}
        \centering
        \includegraphics[width=\linewidth,trim={0 8cm 8cm 0}, clip]{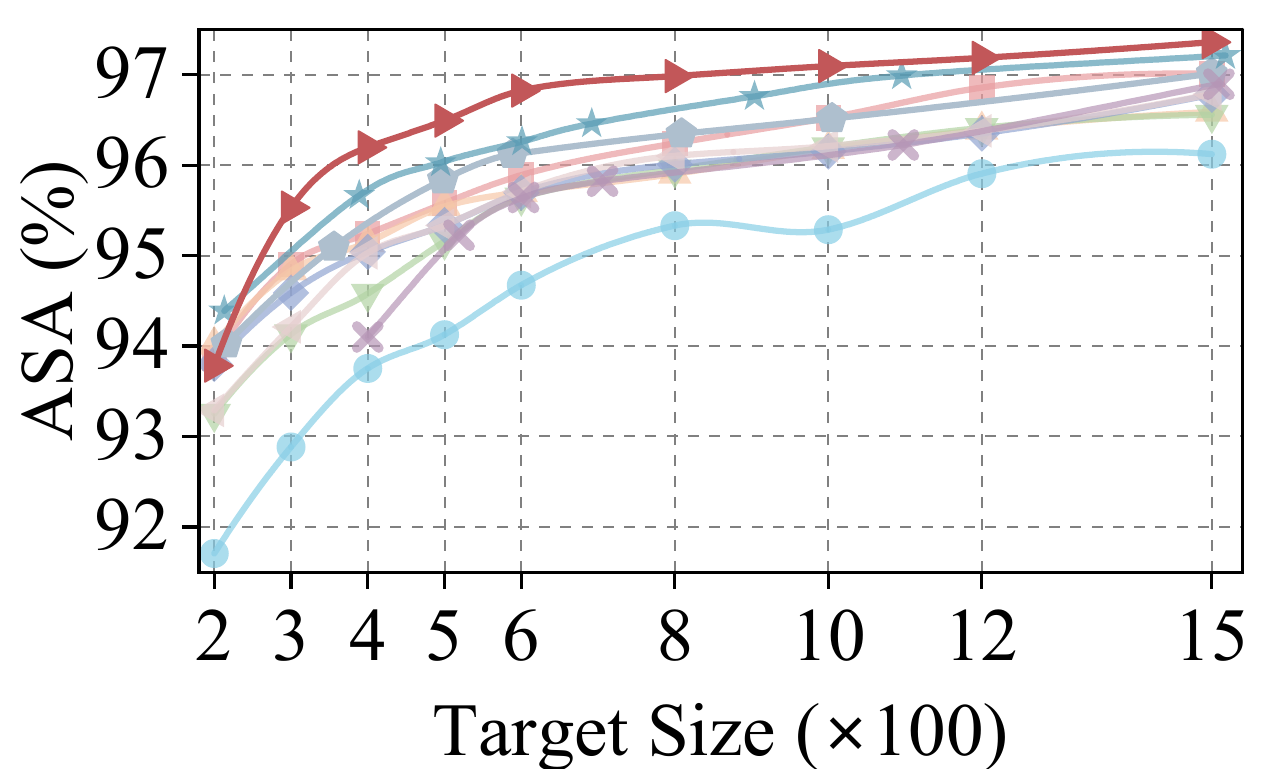}
        % \caption{a}
        \label{fig:ASA_line}
    \end{subfigure}
    \hfill
    \begin{subfigure}{0.245\linewidth}
        \centering
        \includegraphics[width=\linewidth,trim={0 8cm 8cm 0}, clip]{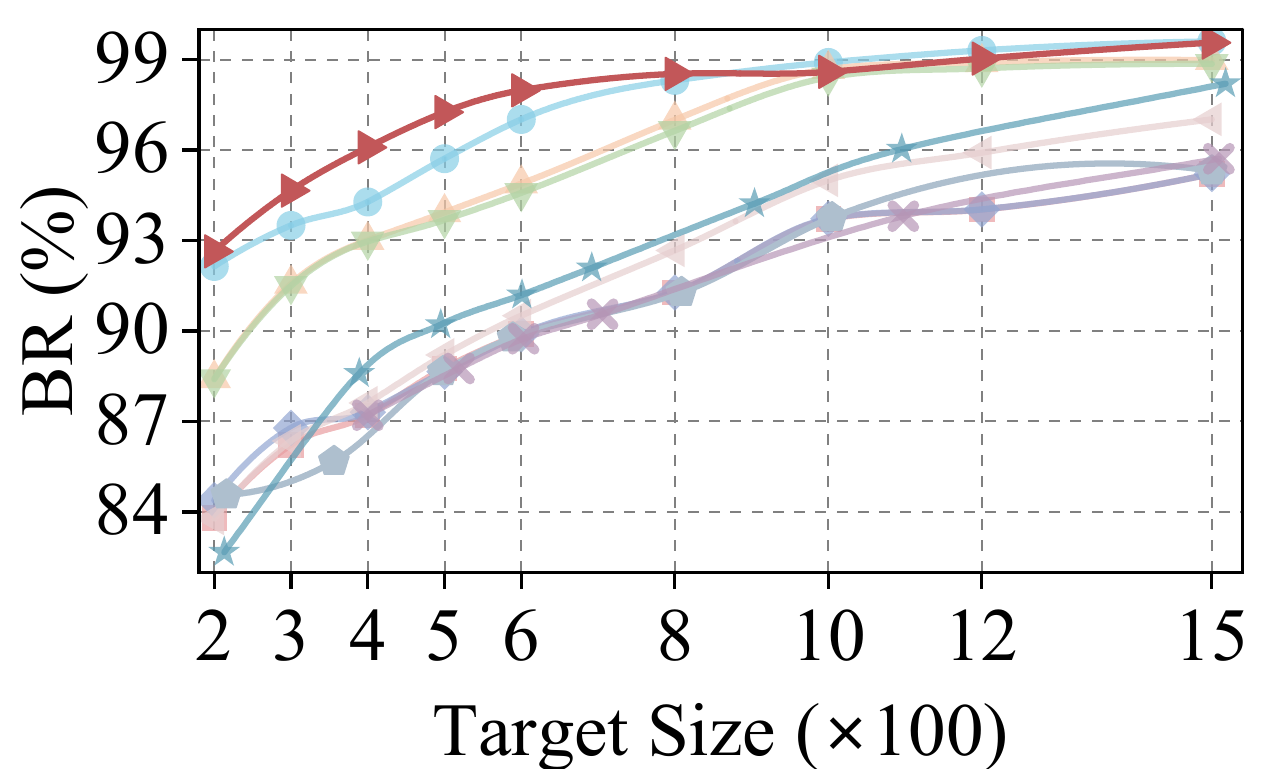}
        % \caption{b}
        \label{fig:BR_line}
    \end{subfigure}
    \hfill
    \begin{subfigure}{0.245\linewidth}
        \centering
        \includegraphics[width=\linewidth,trim={0 8cm 8cm 0}, clip]{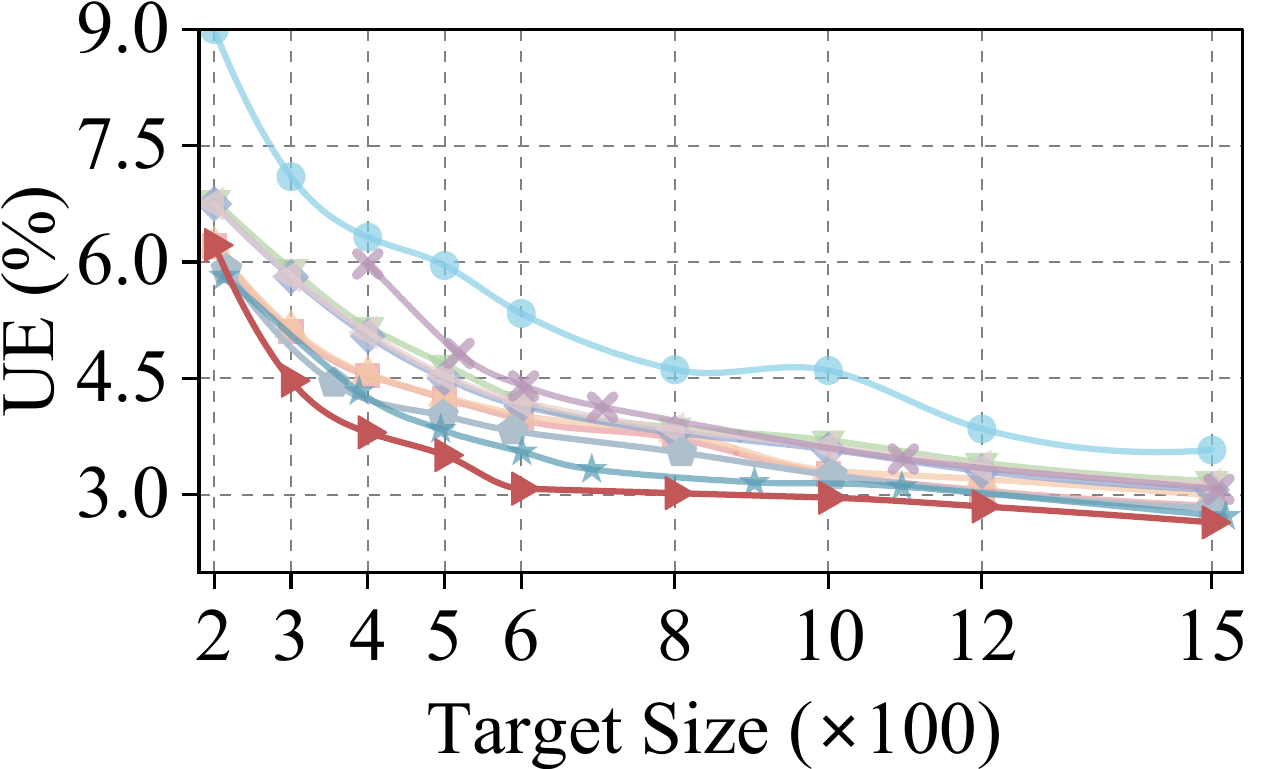}
        % \caption{c}
        \label{fig:UE_line}
    \end{subfigure}
    \hfill
    \begin{subfigure}{0.245\linewidth}
        \centering
        \includegraphics[width=\linewidth,trim={0 8cm 8cm 0}, clip]{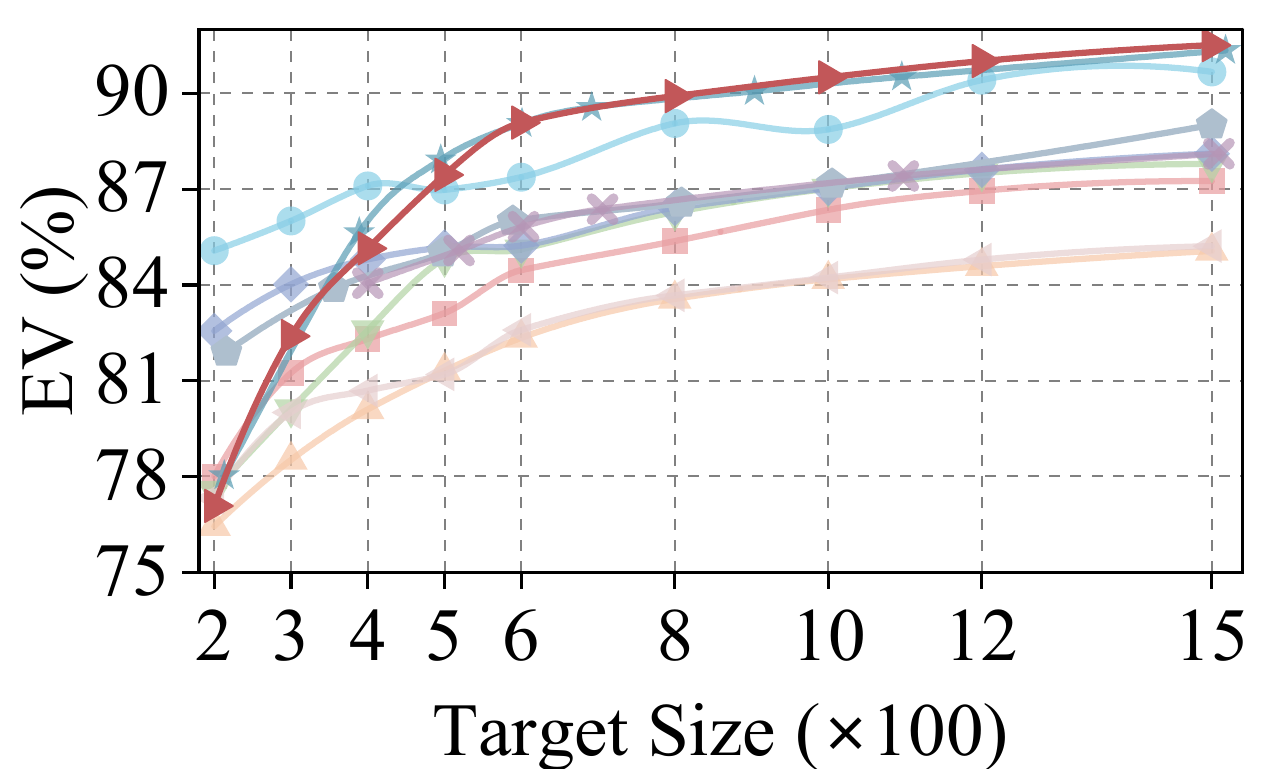}
        % \caption{d}
        \label{fig:EV_line}
    \end{subfigure}
    \vspace{-0.85cm}
    \caption{ASA, BR, UE, and EV Curve of our SIT-HSS compared with other state-of-the-art superpixel segmentation algorithms on the BSDS500 dataset.}\label{fig:line}
    \label{fig_quantitative}
    \vspace{-2.5mm}
\end{figure*}
\subsection{Quantitative Result Comparison (RQ1).}
\begin{figure*}[t]
    \center
    \includegraphics[width=1\textwidth]{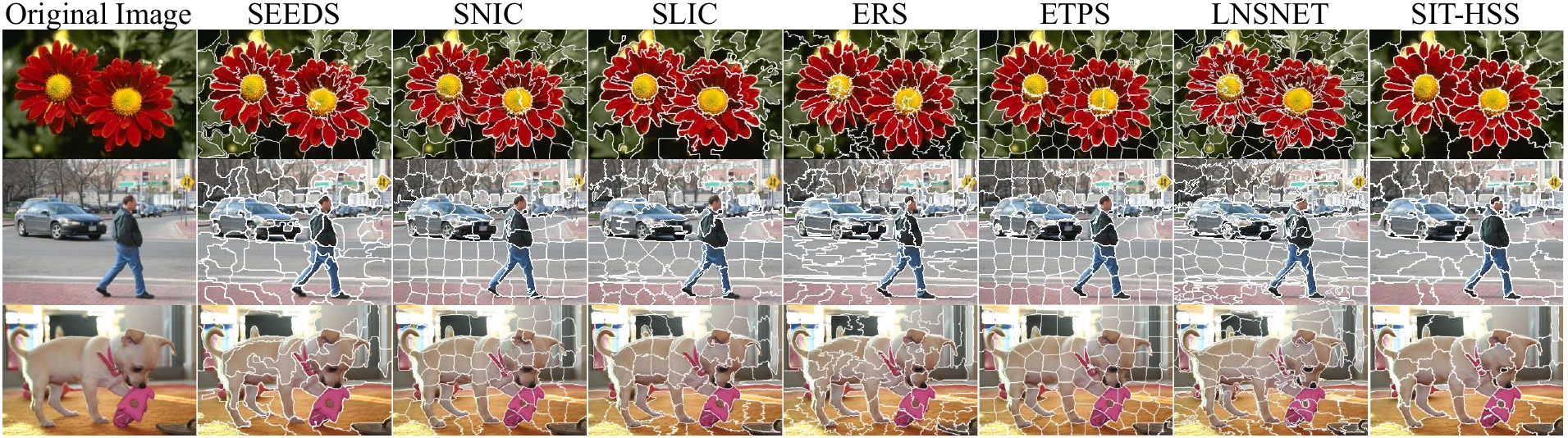}
	\vspace{-1em}
	\caption{Visualization of different superpixel segmentation algorithms on different datasets when $K = 100$.}
	\label{fig_qualitative}
\end{figure*}

Table~\ref{tab_quantitative} shows the quantitative results in both datasets when $K = 600$.
SIT-HSS emerges as the most superior method, outperforming all baselines across both three datasets.
Specifically, for boundary adherence, SIT-HSS achieves the highest performance on the BR metric.
This is owing to SIT-HSS's consideration of the influence of non-adjacent nodes and global information, which allows it to capture more details and improve boundary adherence.
Regarding UE, our method also outperforms other algorithms, clearly demonstrating its capability to reduce under-segmentation errors and excel in weak boundary detection. 
Regarding the consistency of pixels within each superpixel, SIT-HSS shows the best performance on the metric EV on two datasets and the runner-up performance on the third dataset.
The advantage of SIT-HSS on consistency lies in the structural entropy-based cost function, which effectively captures similarities between pixels.
At the same time, the contour evolution-based method ETPS achieves competitive EV values due to its sensitivity to continuous feature variations.
Additionally, SIT-HSS performs well on the ASA metric, indicating that the superpixel generated by our SIT-HSS has a reasonable upper bound for adhering to the object boundaries.
In summary, the SIT-HSS demonstrates exceptional performance across multiple metrics, meeting the universal standards for superpixel segmentation. 
Figure~\ref{fig_quantitative} presents the quantitative comparisons with the SOTA algorithms for different $K$ values in the BSDS500 dataset.
For the results of the SBD and PASCAL-S datasets, please refer to the appendix.

\subsection{Qualitative Result Comparison (RQ2).}
\begin{figure*}[t]
    \center
    \includegraphics[width=1\textwidth]{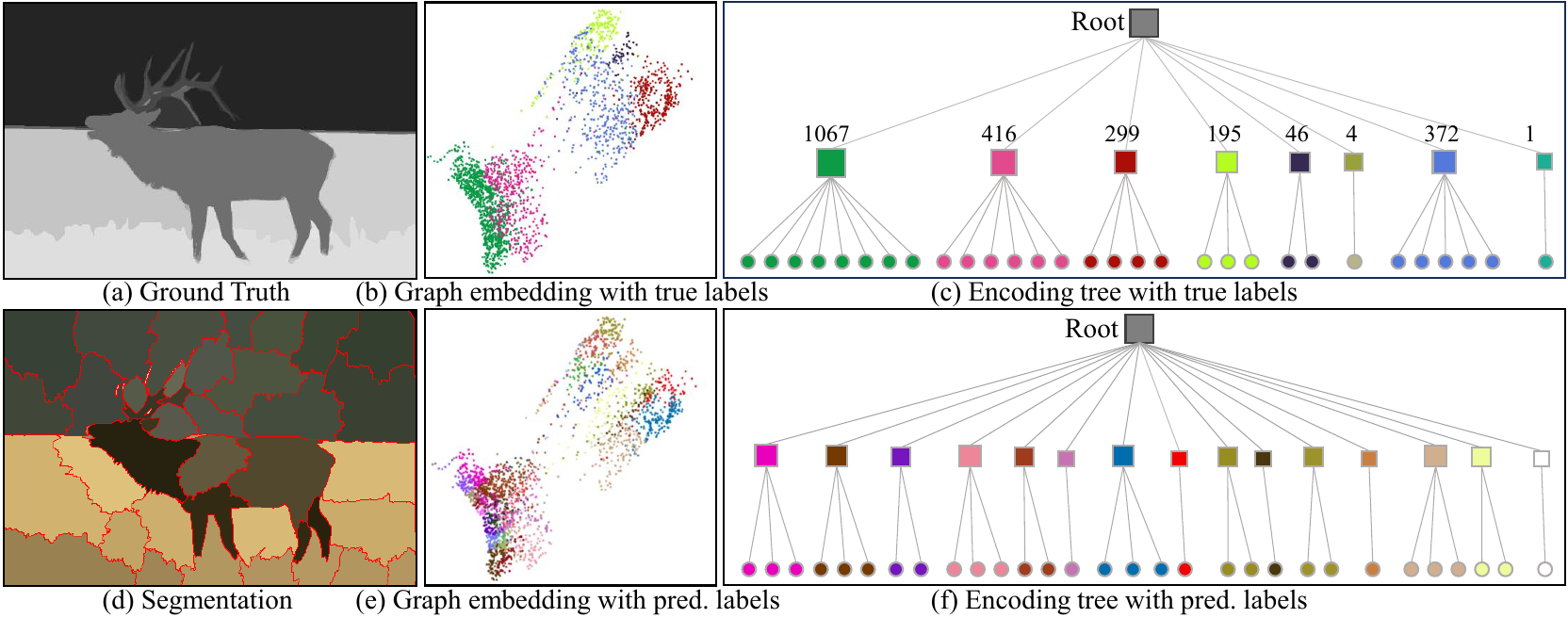}
	\vspace{-0.5cm}
	\caption{Visualization of the effectiveness validation of graph construction and partitioning.}
	\label{fig_springlayout}
    \vspace{-4mm}
\end{figure*}

Fig.~\ref{fig_qualitative} shows a visual comparison of seven representative algorithms on images from three datasets with $K=100$.
Among these algorithms, the segmentations ERS, SEEDS, ETPS, LNSNET, and SIT-HSS exhibit good boundary adherence, with most of the boundaries preserved.
However, ERS and LNSNET generate distorted boundaries, leading to low compactness.
SEEDS and ETPS fail to capture the boundaries of small objects, such as the puppy's eye in the third row.
The segmentations of SLIC and SNIC can produce relatively regular superpixels in a regular shape. 
Nevertheless, they cannot adhere to object boundaries very well.
In all, SIT-HSS achieves good boundary adherence with high compactness.

To further analyze the performance and demonstrate the interpretability of SIT-HSS in superpixel segmentation, we visualize the weighted pixel graph and the structural entropy encoding trees, as shown in Fig.~\ref{fig_springlayout}. 
Specifically, we construct the weighted pixel graph as described in Sect.~\ref{sec_graph_construction} and calculate the positions of the graph nodes using the spring embedder with the Fruchterman-Reingold force-directed algorithm~\cite{springlayout}.
Figs.~\ref{fig_springlayout} (b) and (e) show the graph embedding with ground-truth labels and superpixel labels segmented by SIT-HSS, respectively.
The weighted pixel graph clearly distinguishes the clusters in the ground truth, validating the effectiveness of the SIT-HSS graph construction method.
We construct encoding trees from ground truth labels and superpixel labels and visualize them in Fig.~\ref{fig_springlayout} (c) and (f), respectively.
In these trees, the square tree nodes represent superpixels, and the circular leaf nodes represent pixels.
After segmentation, the superpixel labels closely match the ground truth, indicating that SIT-HSS correctly segmented most objects into superpixels. 
This validates the effectiveness of the graph partitioning process.

\begin{figure*}[t]
    \centering
    \begin{subfigure}{0.49\linewidth}
        \centering
        \includegraphics[width=\linewidth,trim={0.6cm 7.5cm 0 0}, clip]{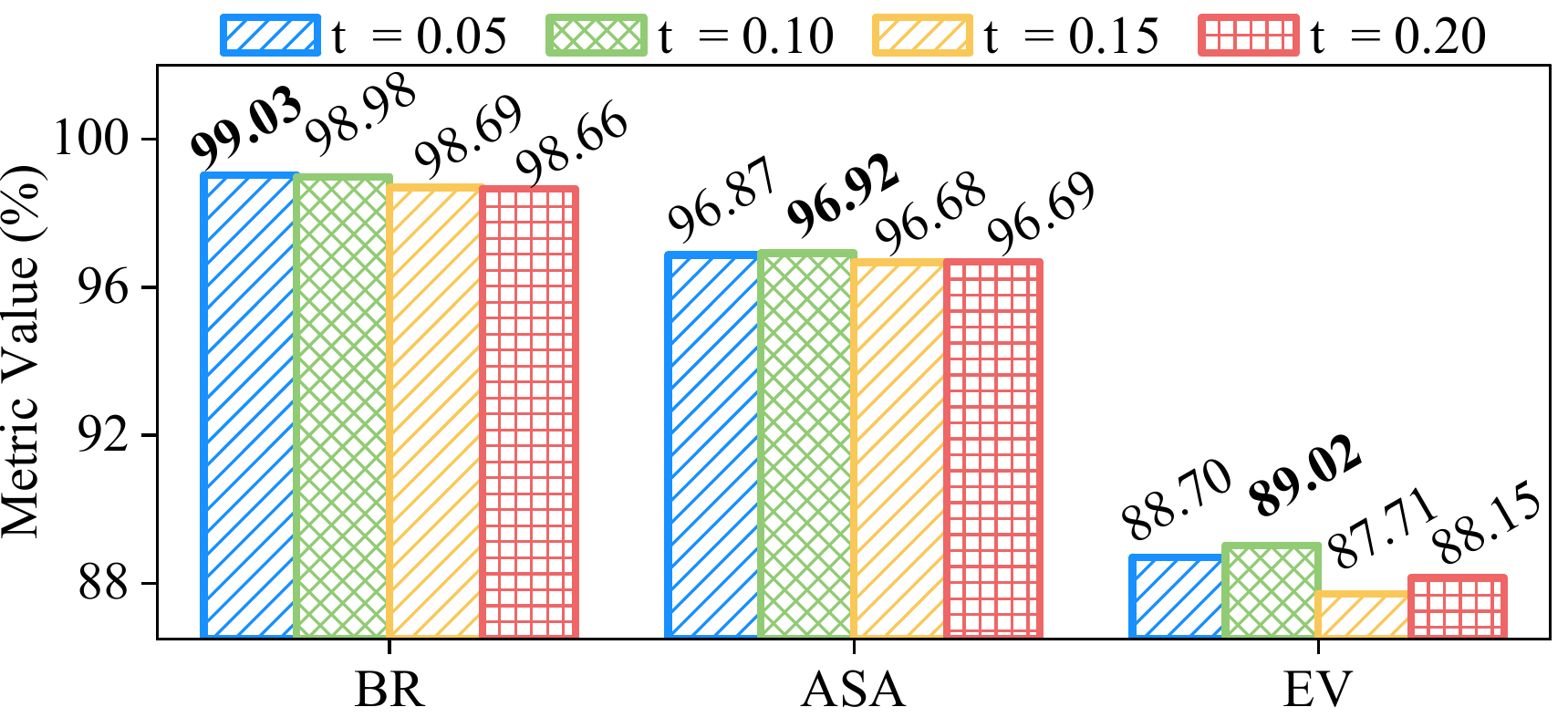}
        \vspace{-0.6cm}
        \caption{Normalization parameter ($t$)}\label{fig:t}
    \end{subfigure}
    \begin{subfigure}{0.49\linewidth}
        \centering
        \includegraphics[width=\linewidth,trim={0.6cm 7.5cm 0 0}, clip]{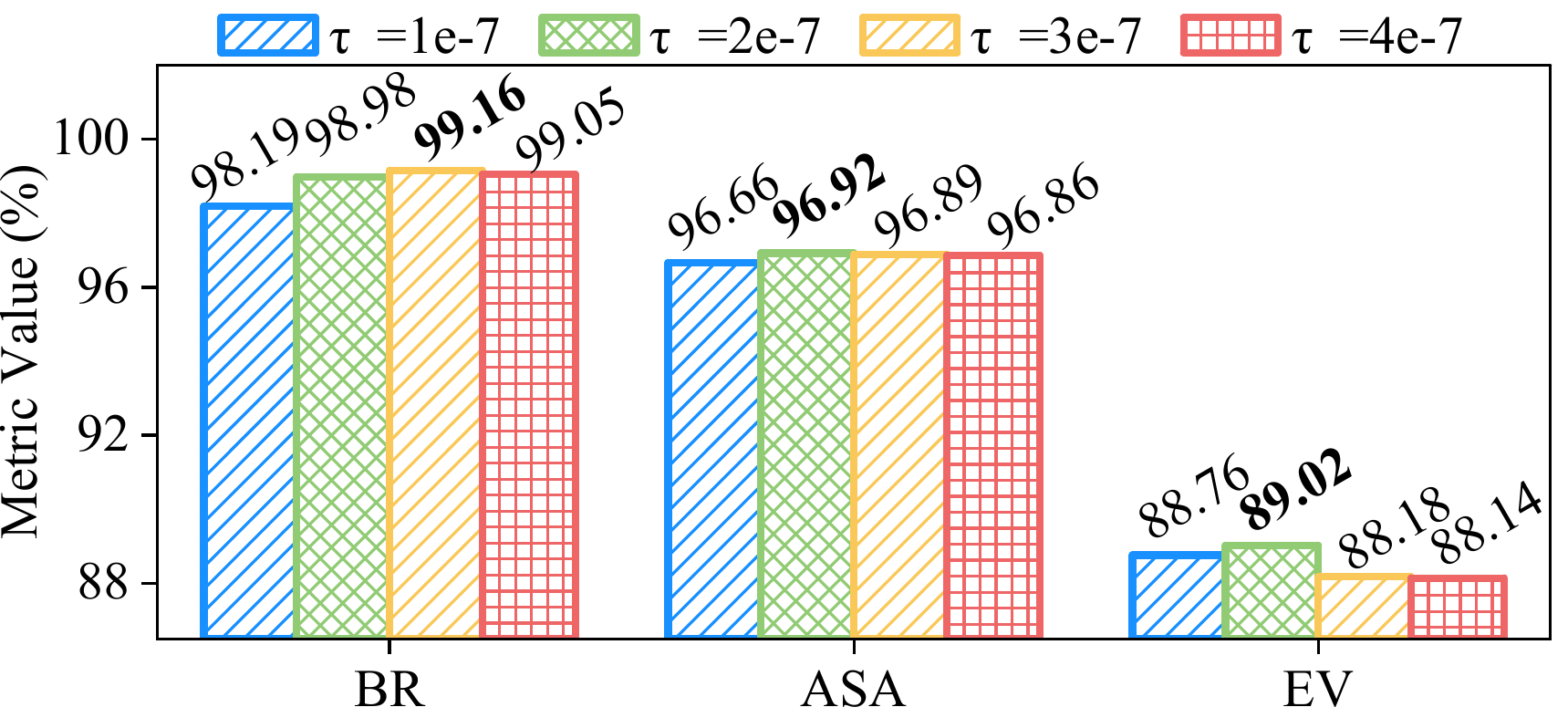}
        \vspace{-0.6cm}
        \caption{SE threshold ($\tau$)}\label{fig:tau}
    \end{subfigure}
    \vspace{-0.2cm}
    \caption{Performance of the SIT-HSS on BSDS500 with different hyperparameter settings.}\label{fig_hyperparameter}
    \vspace{-2mm}
\end{figure*}

\begin{table*}[ht]
    \centering
    \caption{Computational time (s) of different superpixel segmentation algorithms on BSDS500, SBD, and PASCAL-S datasets when $K = 600$.}
    \vspace{-2mm}
    \setlength{\tabcolsep}{4pt}
    \label{tab:performance_metrics}
    \begin{tabular}{l|*{9}{c}c}
        \toprule
        Methods & SLIC & SEEDS & ERS & ETPS & DBSCAN & LSC & SNIC & LNSNET & DRW & \textbf{SIT-HSS} \\
        \midrule
        BSDS500 & 0.138 & 0.102 & 0.361 & 0.124 & 0.034 & 0.159 & 0.139 & 2.860 & 0.888 & 0.141 \\
        SBD & 0.069 & 0.060 & 0.176 & 0.091 & 0.026 & 0.075 & 0.076 & 1.401 & 0.572 & 0.059 \\
        PASCAL-S &0.138  &0.183 &0.574  &0.169  &0.042  &0.167  &0.213  &4.045 &0.980 &0.154  \\
        \midrule
        Average &0.115  &0.115  &0.370  &0.128  &0.034  &0.134  &0.143  &2.769  &0.813  &0.118 \\
        % \midrule
        % Complexity &\(O(N)\)&\(O(N)\)&\(O(N^2\log N)\) &\(O(N)\)&\(O(N)\)&\(O(N)\)&\(O(N)\)&\(O(N)\)&\(O(N)\)&\(O(E)\)\\
        \bottomrule
    \end{tabular}
    \label{tab_efficiency}
 \vspace{-3mm}
\end{table*}

\subsection{Hyperparameter Study (RQ3).}
As shown in Fig.~\ref{fig_hyperparameter}, we conduct the hyperparameter experiments on the BSDS500 dataset to investigate the impact of the weight normalization parameter \( t \) (which controls the sensitivity of pixel similarity to distance in Eq.~\ref{eq_3.4}) and the SE threshold \( \tau \) (used to determine the search radius \( r \)). 
Fig.~\ref{fig:t} illustrates that when \( t \) is small (e.g. 0.05), the weight normalization function changes rapidly, leading to significant weight differences even with minor changes in the distance of the pixels.
This sensitivity to pixel variation increases the BR value but also makes the segmentation result more susceptible to noise, reducing ASA and EV.
In contrast, when \( t \) is large (e.g. 0.20), the weight function becomes less sensitive to distance, resulting in smoother segmentation.
This reduces over-segmentation but potentially causes a loss of boundary details and slightly lowers BR value.
When \( t = 0.10 \), ASA and EV reach their peak values, indicating that a moderate \( t \) balances segmentation accuracy and boundary preservation well. 
Furthermore, Fig.~\ref{fig:tau} shows that when the SE threshold \( \tau \) is small (e.g. 1e-7), the selected neighborhood radius \( r \) is large, making the graph overly dense, potentially containing unimportant information or even noise.
When \( \tau \) is large (e.g., 4e-7), the radius of the neighbourhood becomes small, resulting in an overly sparse graph structure that retains insufficient information. 
An appropriate value of \( \tau \) helps to prevent an overly complex graph structure while retaining valuable information in the graph. 
For hyperparameter studies on other datasets, please refer to the Appendix.

\subsection{Efficiency (RQ4).}
Table~\ref{tab_efficiency} reports the average processing time of each method per image.
SIT-HSS achieves top-tier efficiency among these methods.
Compared to the fastest method, DBSCAN, SIT-HSS requires only an additional 0.084 seconds per image, demonstrating a comparable speed.
In particular, SIT-HSS speeds up 3 to 10 times faster than other graph-based methods, such as ERS and DRW.
Compared to the deep learning-based LNSNET method, which requires an average of 2.86 seconds, SIT-HSS significantly reduces processing time. 
Although SLIC, SEEDS, and DBSCAN perform marginally better in speed, SIT-HSS achieves a better balance between segmentation accuracy and efficiency. 
Furthermore, by adopting more powerful GPUs and advanced hardware, the processing time of SIT-HSS can be further reduced, making it easily scalable for large-scale image data processing.
\section{RELATED WORK}\label{sec: RelatedWork}
In the last two decades, superpixel segmentation has made significant progress in terms of quality and efficiency, and numerous approaches have emerged.
Among them, prevalent approaches include graph-based, cluster-based, and deep learning-based methods.
Graph-based methods model the segmentation problem using graph theory and generate superpixels by minimizing a cost function defined on the graph~\cite{shi2000normalized, ERS, DRW}.
Clustering-based methods group pixels and iteratively refine them until certain convergence criteria are met~\cite{SLIC, SEEDS, SCSC}.
More recently, several deep learning-based methods \cite{FCN, Alnet, LNSNet} have been introduced, utilizing the high representation power of neural networks.
However, their lack of interpretability and high computational complexity limit their range of applications. 
 
Hierarchical superpixel segmentation highlights the importance of multiresolution representations.
These approaches generate a set of superpixel segmentations at different scales, where a coarser superpixel is composed of finer ones.
LASH~\cite{LASH} agglomerates pixels into superpixel hierarchy based on a similarity function by reinforcement learning.
SH~\cite{SH} constructs the superpixel hierarchy using minimum spanning trees (MST). 
CRTrees~\cite{CRTrees} extend SH's undirected MST into directed structures based on 1-NN graph theory, allowing for the automatic determination of the number of superpixels at each hierarchy.
HHTS~\cite{chang2024hierarchical} iteratively partitions the segments top-down by thresholding.
SIT-HSS achieves hierarchical agglomerative superpixel segmentation by hierarchical partitioning on pixel graphs, which retains adequate information through 1D SE maximization.
\section{CONCLUSION}\label{sec: Conclusion}
This paper introduces SIT-HSS, a novel hierarchical superpixel segmentation method via structural information theory.
SIT-HSS achieves superpixel segmentation via hierarchical graph partitioning on pixel graphs based on the structural entropy.
Unlike existing graph-based methods that consider only the relations of directly adjacent pixels, SIT-HSS constructs graphs by incrementally exploring the pixel neighborhood.
This graph construction strategy maximizes the retention of graph information while avoiding an overly complex graph structure by maximizing 1D SE.
Based on the pixel graph, SIT-HSS iteratively merges graph nodes into a hierarchical graph partitioning by 2D SE minimization, forming a hierarchical superpixel segmentation.
Extensive experiments demonstrate that SIT-HSS achieves SOTA performance with top-tier efficiency.

\section*{Acknowledgments}
This work is supported by the NSFC through grants 62322202, 62441612, and 62432006, the Shijiazhuang Science and Technology Plan Project through grant 231130459A, and the Guangdong Basic and Applied Basic Research Foundation through grant 2023B1515120020.

\bibliographystyle{siamplain}
\bibliography{references}
\clearpage
\section*{Appendix}
\subsection{Quantitative Result Comparison.}
Fig.~\ref{fig_quantitative_SBD} and~\ref{fig_quantitative_S} illustrate the quantitative result of ASA, BR, UE, and EV of nine advanced superpixel segmentation algorithms and our SIT-HSS on SBD and PASCAL-S datasets.
SIT-HSS demonstrates strong performance in multiple metrics. 
Its improvement in the ASA metric highlights the algorithm's ability to maintain similarity to the target, particularly notable at high superpixel counts. 
In contrast, while other algorithms perform well in certain scenarios, they generally fall short compared to SIT-HSS. 
In terms of the BR and UE metrics, SIT-HSS also excels, effectively capturing target boundaries and reducing under-segmentation errors, thereby enhancing the accuracy of the segmentation results. 
Regarding the EV metric, SIT-HSS maintains a high level of feature retention capability and exhibits excellent structural consistency, further affirming its advantages in superpixel segmentation tasks.

\subsection{Hyperparameter Study.} \label{appendix_parameter}
As \( t \) and \( \tau \) vary, we examine the changes in the performance of SIT-HSS on SBD and PASCAL-S datasets. 
The results are presented in Fig.~\ref{fig:hyperparameter_SBD} and~\ref{fig:hyperparameter_PASCAL-S}.

Weight normalization parameter \( t \): The selected values of \( t \) are 0.05, 0.10, 0.15, and 0.20. 
Figs.~\ref{fig:S_t} and~\ref{fig:P_t} indicate that SIT-HSS performs best when \( t = 0.1 \). 
Smaller \( t \) values result in a decrease in the overall precision of the segmentation, while larger \( t \) values lead to less refined boundaries.

Structural entropy threshold \( \tau \): 
The selection of the value \( \tau \) is based on the characteristics of the datasets and the model sensitivity analysis. 
For the SBD dataset, where the image complexity is relatively lower, larger thresholds help reduce unnecessary noise.
Therefore, we set the values of \( \tau \) at $1e-6$, $2e-6$, $3e-6$, and $4e-6$.
As shown in Fig.~\ref{fig:S_tau}, when \( \tau \) is set to $1e-6$, SIT-HSS performs best on SBD. 
In contrast, as illustrated in Fig.~\ref{fig:P_tau}, when \( \tau \) is set to $2e-7$, SIT-HSS achieves optimal performance on PASCAL-S. 
Smaller \( \tau \) values result in lower boundary adhesion, while larger \( \tau \) values cause the model to focus too much on the boundaries, resulting in a slight decrease in the consistency of the region and the overall quality of segmentation.

\subsection{Convergence Analysis.}
The proposed algorithm is capable of rapidly converging in a few iterations, achieving accurate superpixel segmentation. 
To validate this, we randomly select an image from each dataset and record the changes in the number of superpixels and the number of edges between them during each iteration. 
As shown in Fig.~\ref{fig:iteration}, as iterations progress, both the number of superpixels and the number of edges exhibit an exponential decline.
 On average, by the seventh iteration, the number of superpixels approaches the target segmentation size, and by the 10th to 12th iteration, it precisely reaches the target size of 600.
Note that the algorithm's time complexity is linearly related to the number of edges between superpixels. 
As the number of edges decreases, the computation time required for subsequent iterations also decreases accordingly.
\setcounter{figure}{8}
\begin{figure}[t]
\vspace{-5mm}
    \centering
    \begin{subfigure}{0.9\linewidth}
        \centering
        \includegraphics[width=\linewidth]{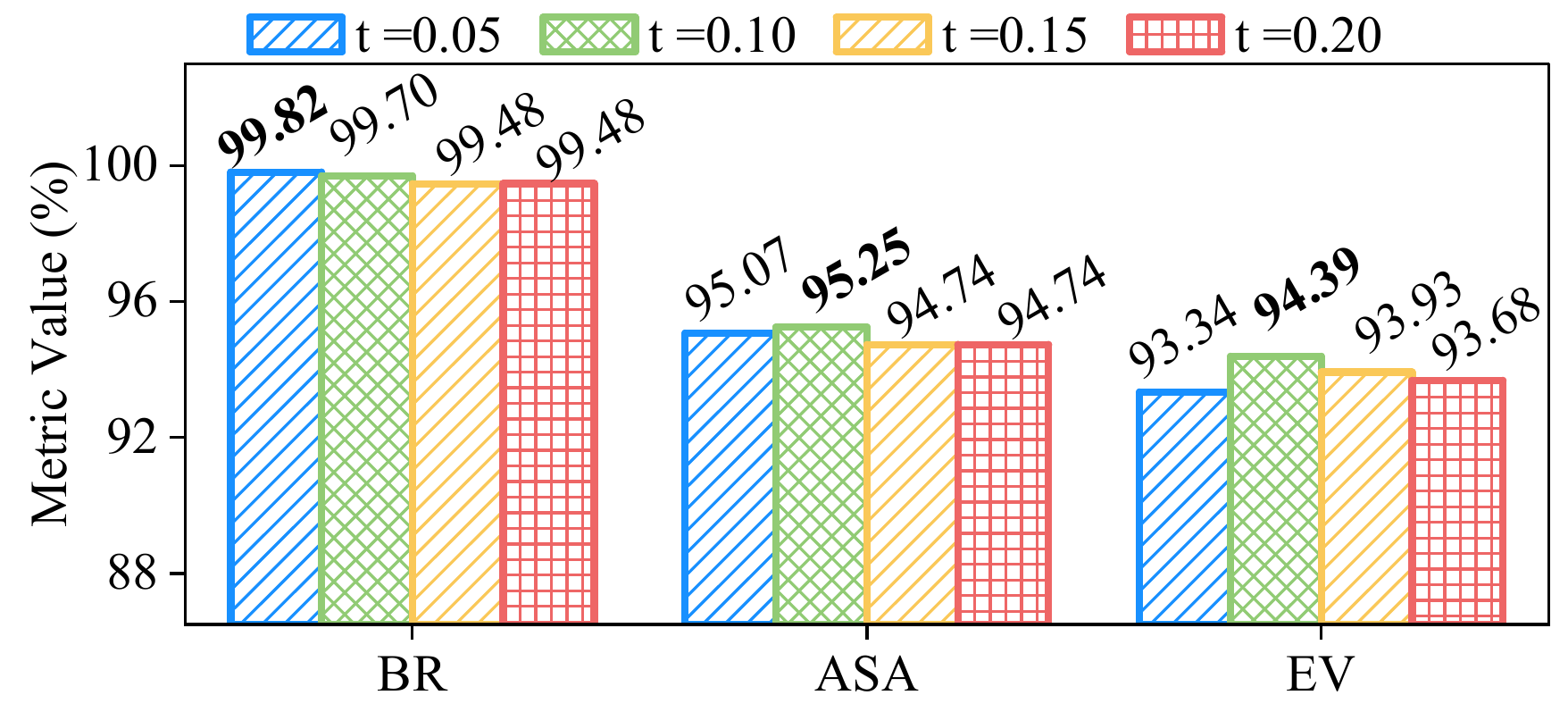}
         \vspace{-6mm}
        \caption{Normalization parameter ($t$)}\label{fig:S_t}
    \end{subfigure}
    \vspace{0.5cm} 
    \begin{subfigure}{0.9\linewidth}
        \centering
        \includegraphics[width=\linewidth]{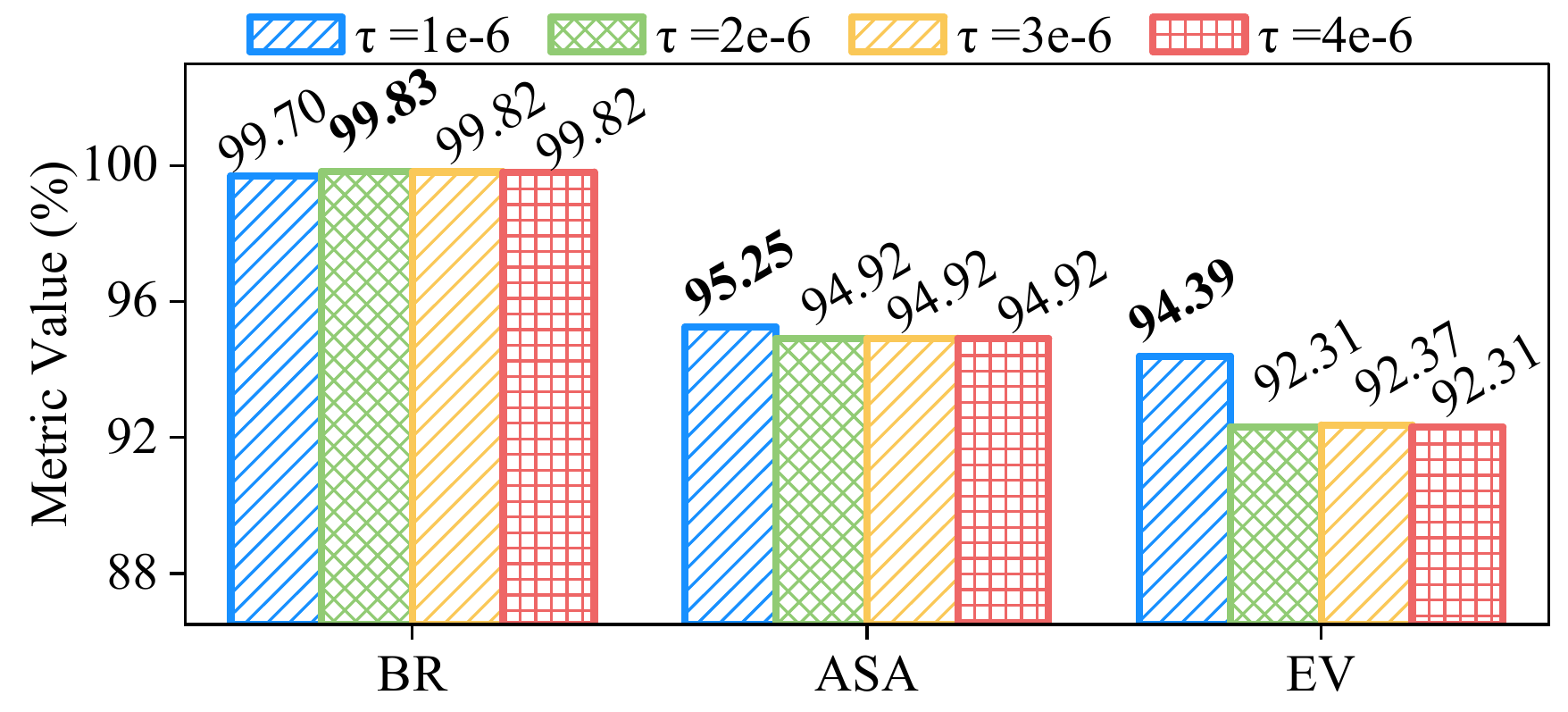}
          \vspace{-6mm}
        \caption{SE threshold ($\tau$)}\label{fig:S_tau}
    \end{subfigure}
    \vspace{-7mm}
    \caption{Performance of the SIT-HSS on SBD with different hyperparameter settings.}\label{fig:hyperparameter_SBD}
\end{figure}
\begin{figure}[t!]
    \centering
    \begin{subfigure}{0.9\linewidth}
        \centering
        \includegraphics[width=\linewidth]{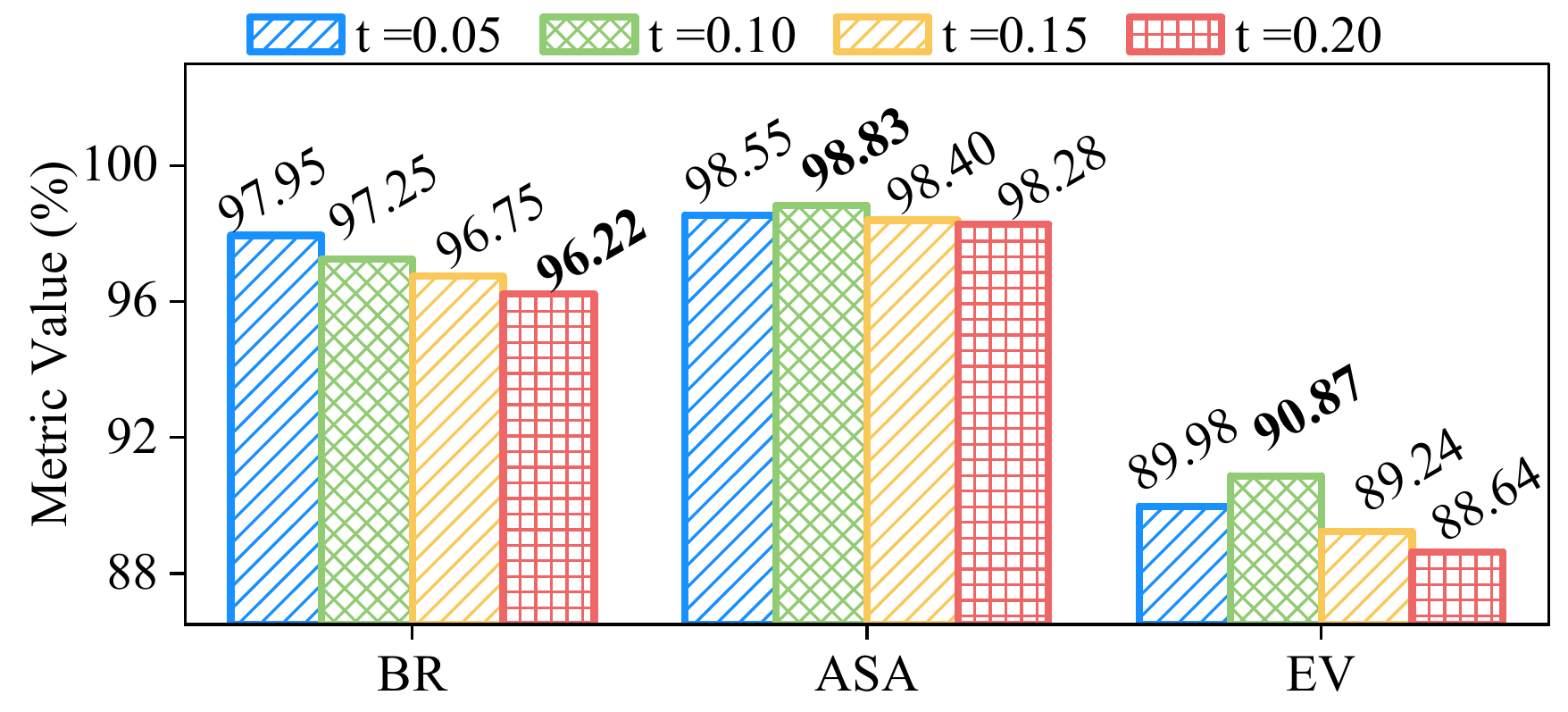}
         \vspace{-6mm}
        \caption{Normalization parameter ($t$)}\label{fig:P_t}
    \end{subfigure}
    \vspace{0.5cm} 
    \begin{subfigure}{0.9\linewidth}
        \centering
        \includegraphics[width=\linewidth]{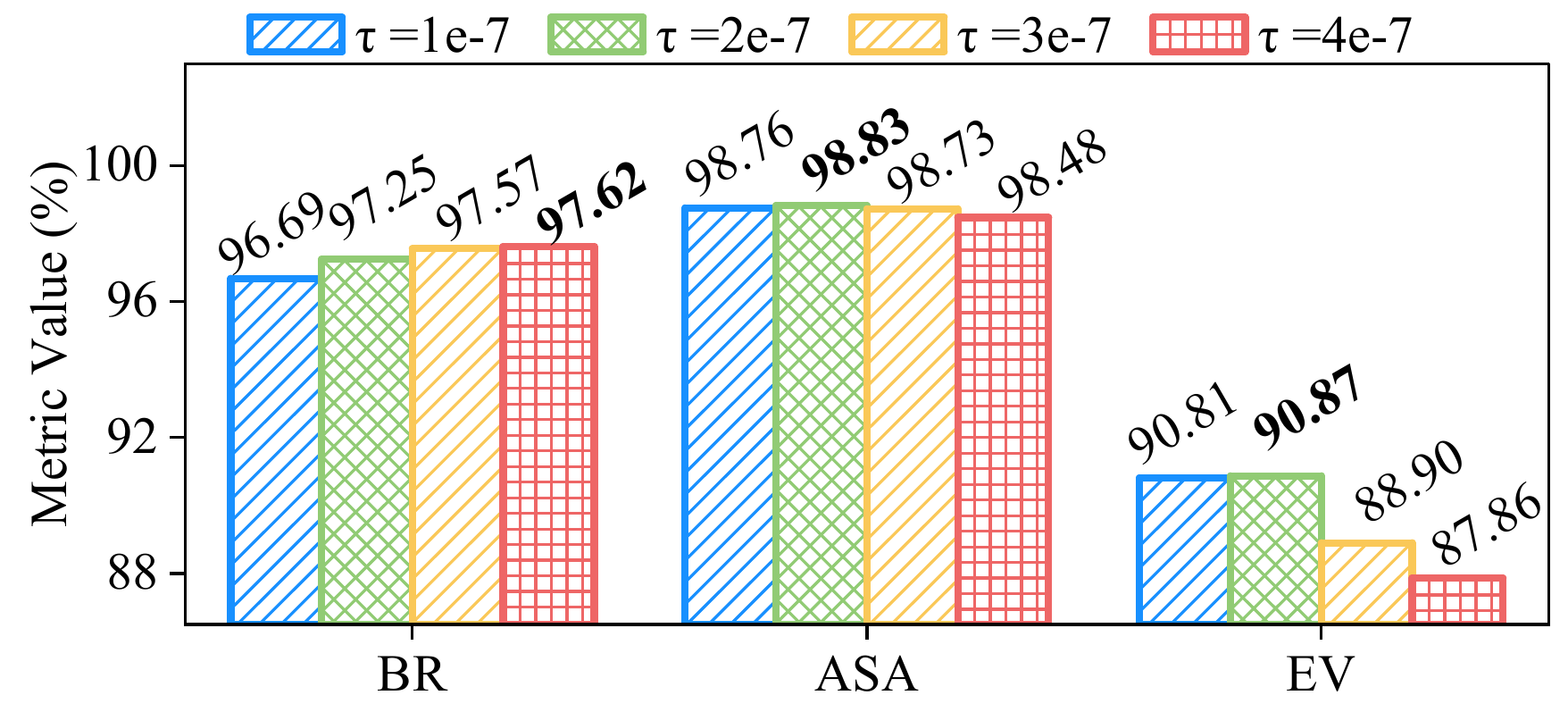}
         \vspace{-6mm}
        \caption{SE threshold ($\tau$)}\label{fig:P_tau}
    \end{subfigure}
    \vspace{-7mm}
    \caption{Performance of the SIT-HSS on PASCAL-S with different hyperparameter settings.}\label{fig:hyperparameter_PASCAL-S}
\end{figure}

\setcounter{figure}{6}
\begin{figure*}[t]
    \centering
    \begin{subfigure}{1\linewidth}
        \centering
    \includegraphics[width=\linewidth,trim={0 1.5cm 0.4cm 1.5cm}, clip]{figures/line_legend.pdf}
    \end{subfigure}
    \begin{subfigure}{0.245\linewidth}
        \centering
        \includegraphics[width=\linewidth,trim={0 8cm 8.5cm 0}, clip]{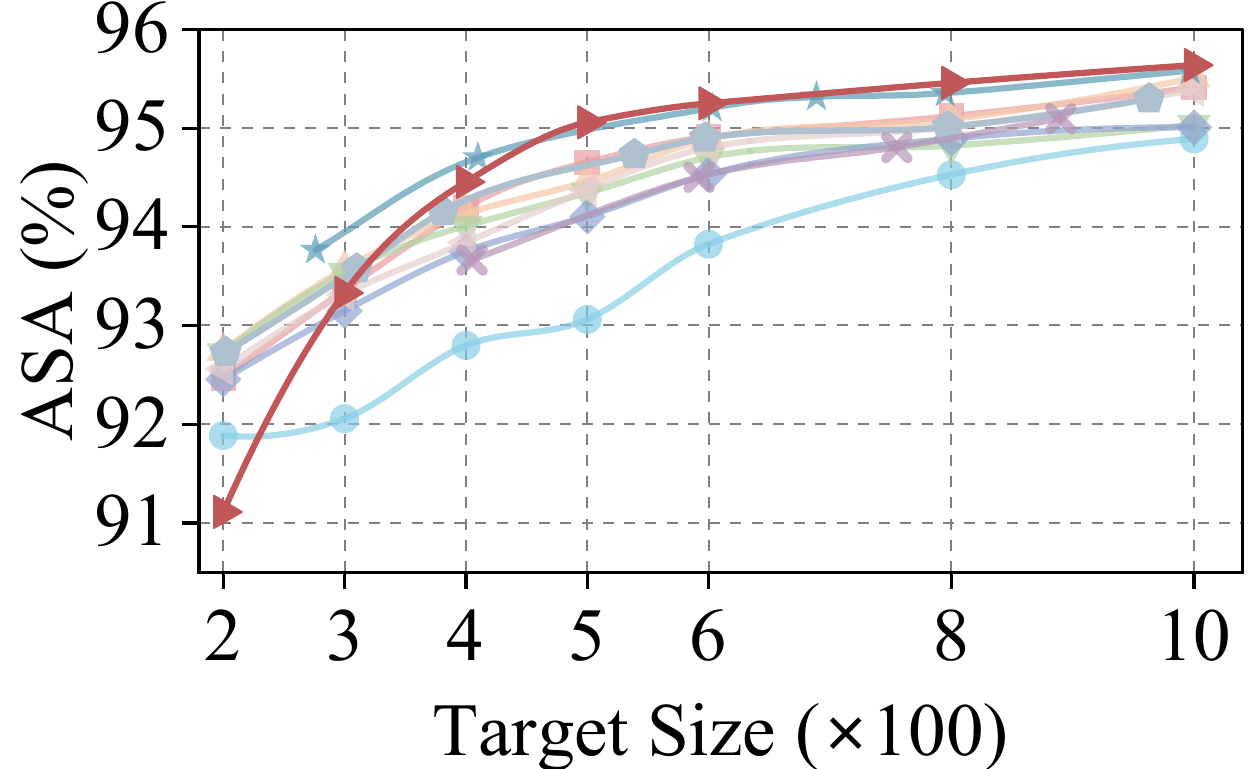}
    \end{subfigure}
    \hfill
    \begin{subfigure}{0.245\linewidth}
        \centering
        \includegraphics[width=\linewidth,trim={0 8cm 8.5cm 0}, clip]{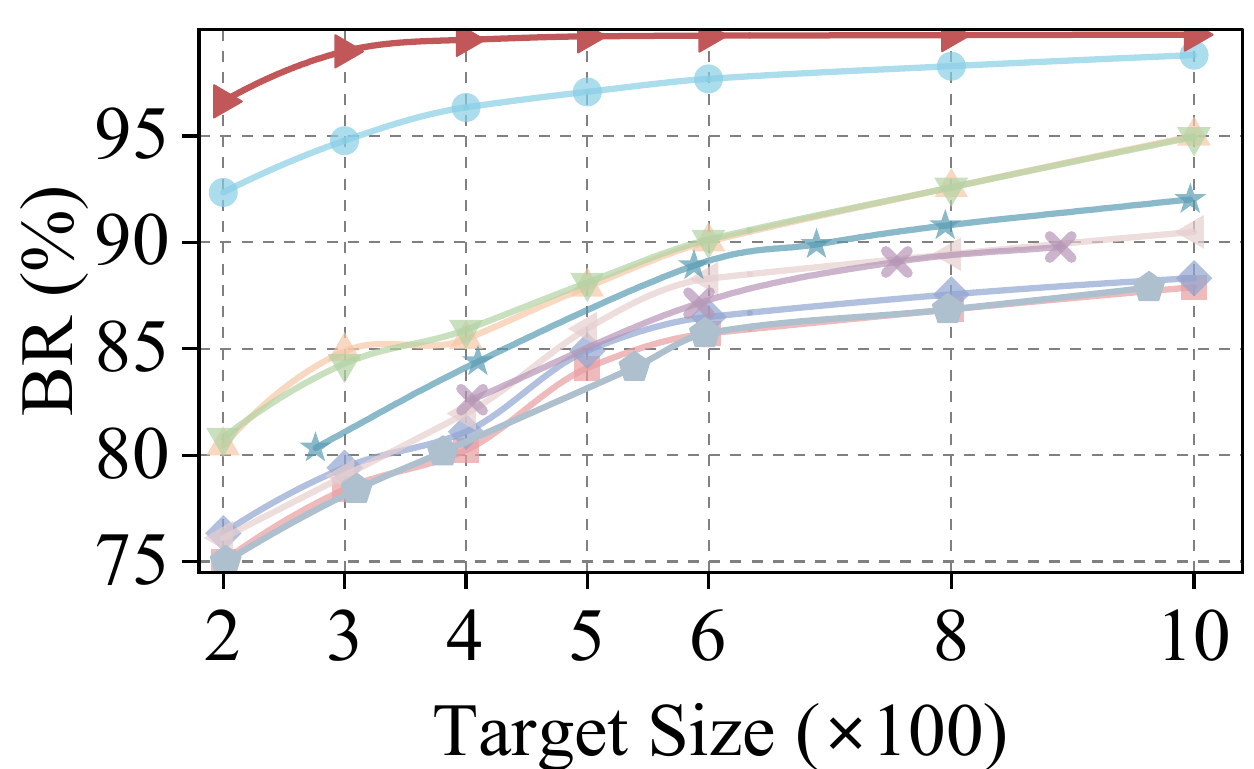}
        % \caption{b}
    \end{subfigure}
    \hfill
    \begin{subfigure}{0.245\linewidth}
        \centering
        \includegraphics[width=\linewidth,trim={0 8cm 8.5cm 0}, clip]{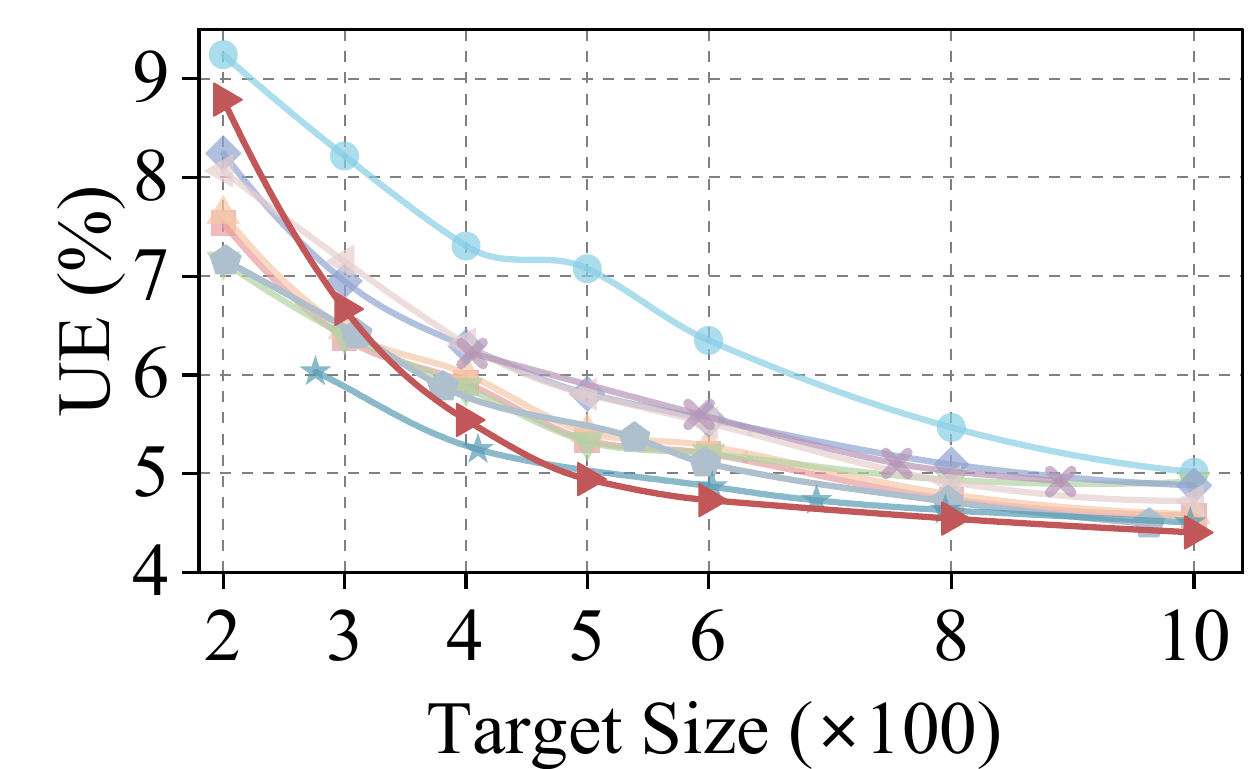}
    \end{subfigure}
    \hfill
    \begin{subfigure}{0.245\linewidth}
        \centering
        \includegraphics[width=\linewidth,trim={0 8cm 8.5cm 0}, clip]{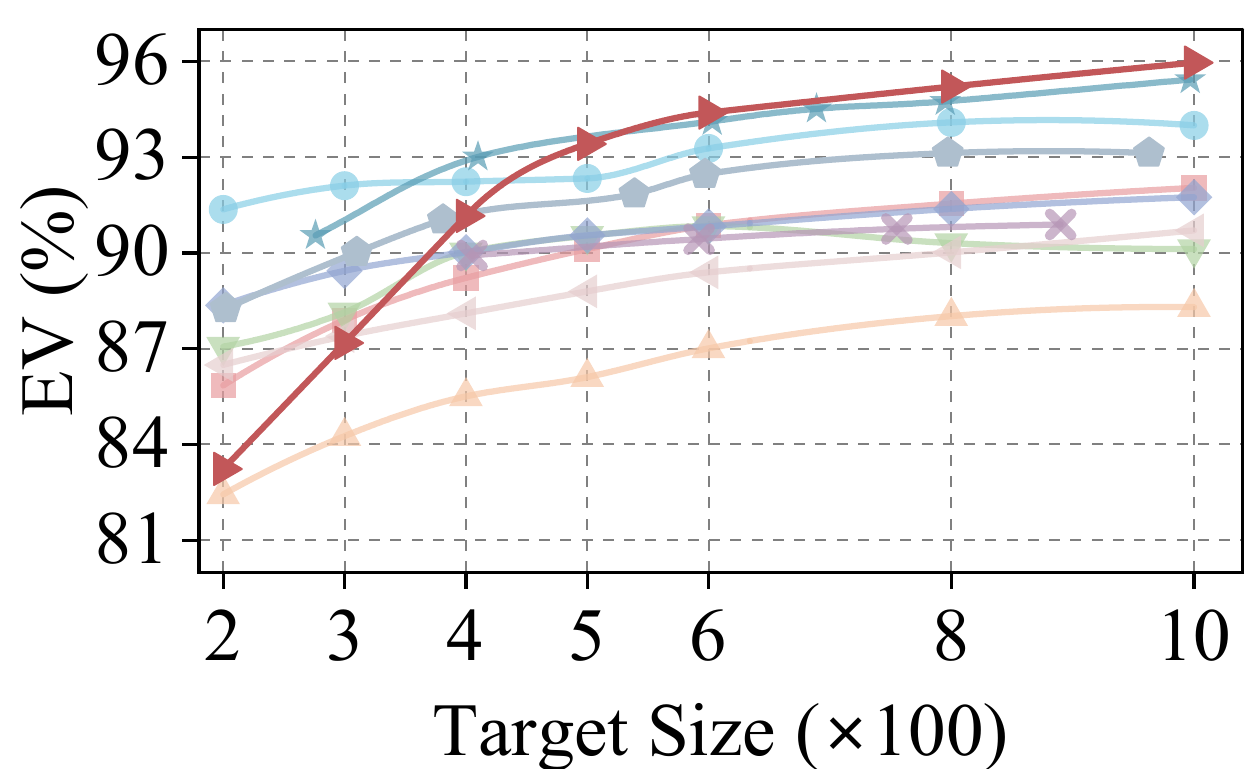}
    \end{subfigure}
    % \vspace{-7mm}
    \caption{ASA, BR, UE, and EV Curve of our SIT-HSS compared with other state-of-the-art superpixel segmentation algorithms on the SBD dataset.}
    \label{fig_quantitative_SBD}
\end{figure*}
\begin{figure*}[t]
    \centering
    \begin{subfigure}{1\linewidth}
        \centering
    \includegraphics[width=\linewidth,trim={0 1.5cm 0.4cm 1.5cm}, clip]{figures/line_legend.pdf}
    \end{subfigure}
    \begin{subfigure}{0.245\linewidth}
        \centering
        \includegraphics[width=\linewidth,trim={0 8cm 8.5cm 0}, clip]{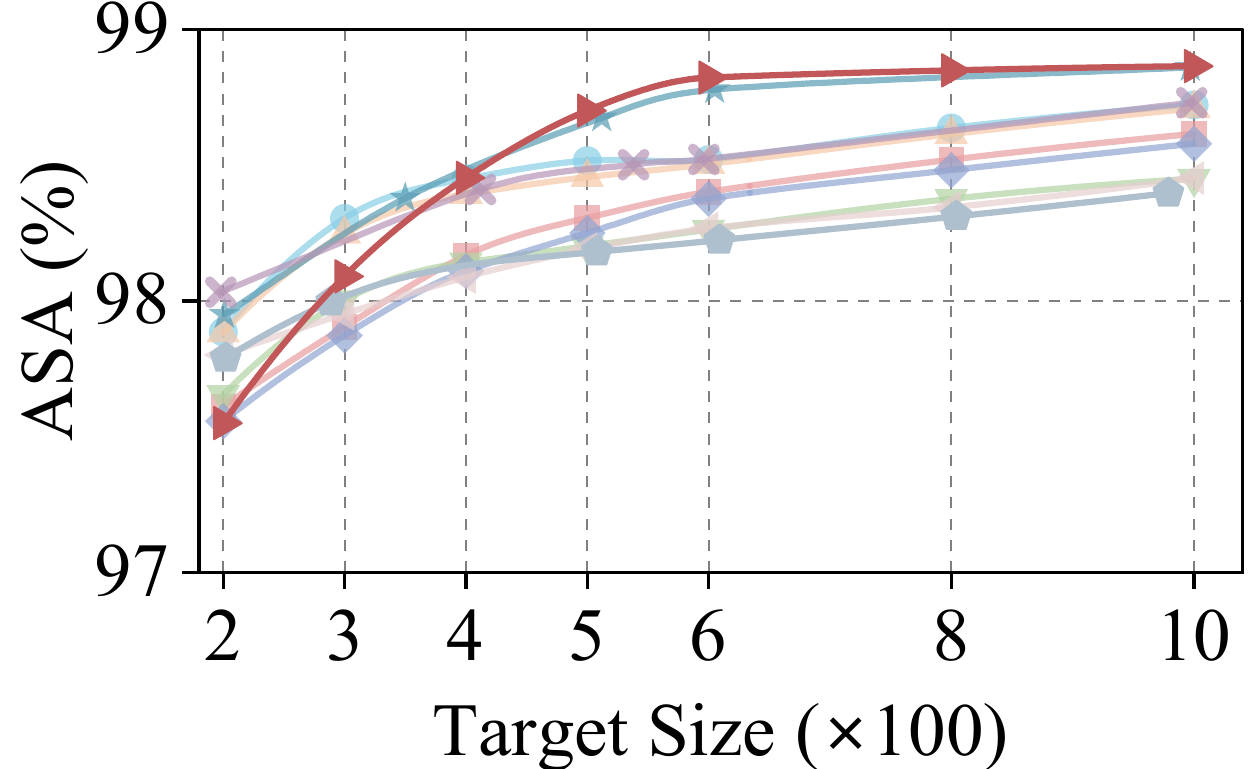}
    \end{subfigure}
    \hfill
    \begin{subfigure}{0.245\linewidth}
        \centering
        \includegraphics[width=\linewidth,trim={0 8cm 8.5cm 0}, clip]{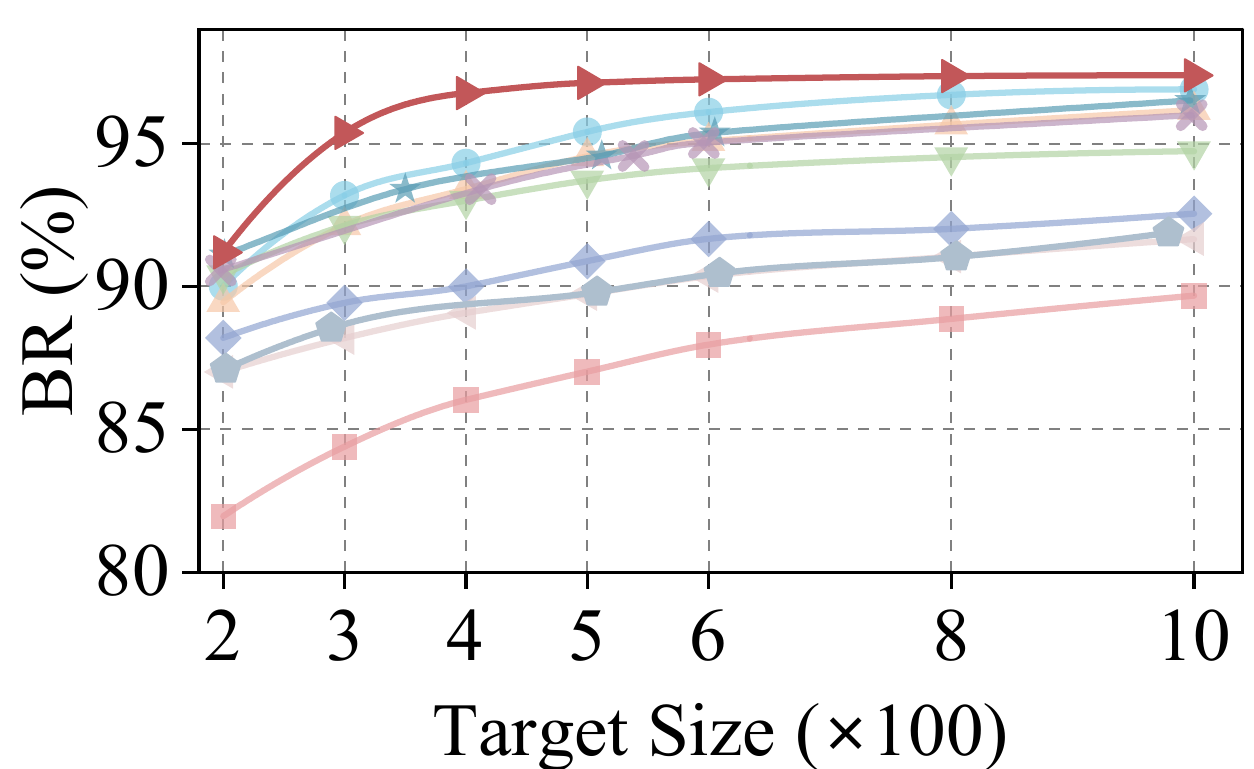}

    \end{subfigure}
    \hfill
    \begin{subfigure}{0.245\linewidth}
        \centering
        \includegraphics[width=\linewidth,trim={0 8cm 8.5cm 0}, clip]{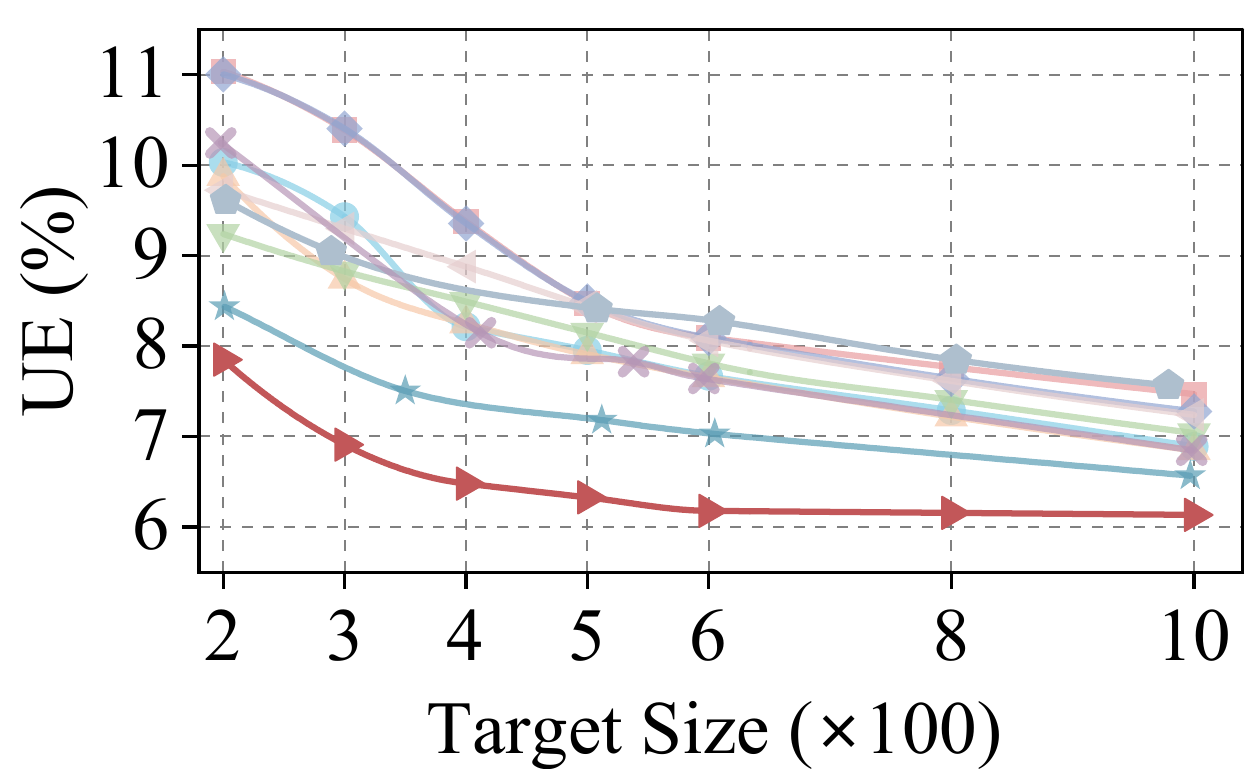}

    \end{subfigure}
    \hfill
    \begin{subfigure}{0.245\linewidth}
        \centering
        \includegraphics[width=\linewidth,trim={0 8cm 8.5cm 0}, clip]{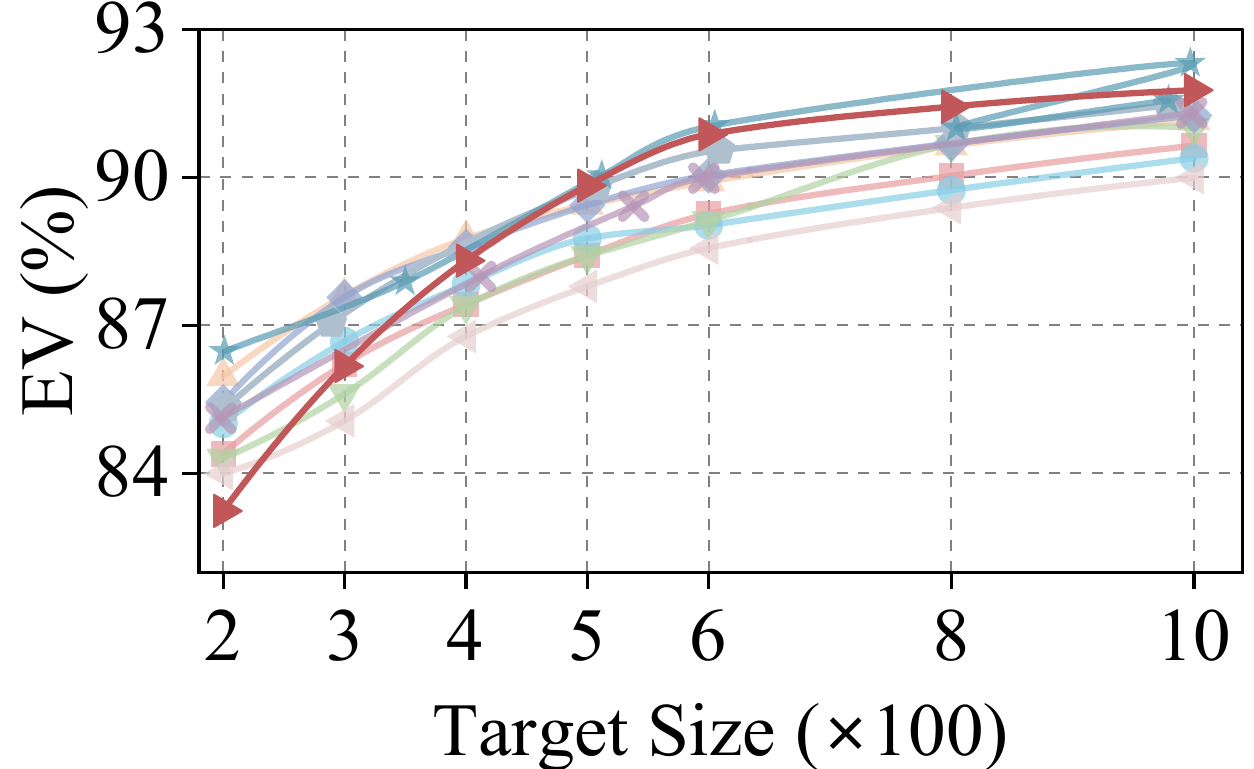}
    \end{subfigure}
    \caption{ASA, BR, UE, and EV Curve of our SIT-HSS compared with other state-of-the-art superpixel segmentation algorithms on the PASCAL-S dataset.}
    \label{fig_quantitative_S}
\end{figure*}

\setcounter{figure}{10}
\begin{figure*}[t]
    \centering
    \begin{subfigure}{0.325\linewidth}
        \centering
        \includegraphics[width=\linewidth,trim={0 1cm 0 0}, clip]{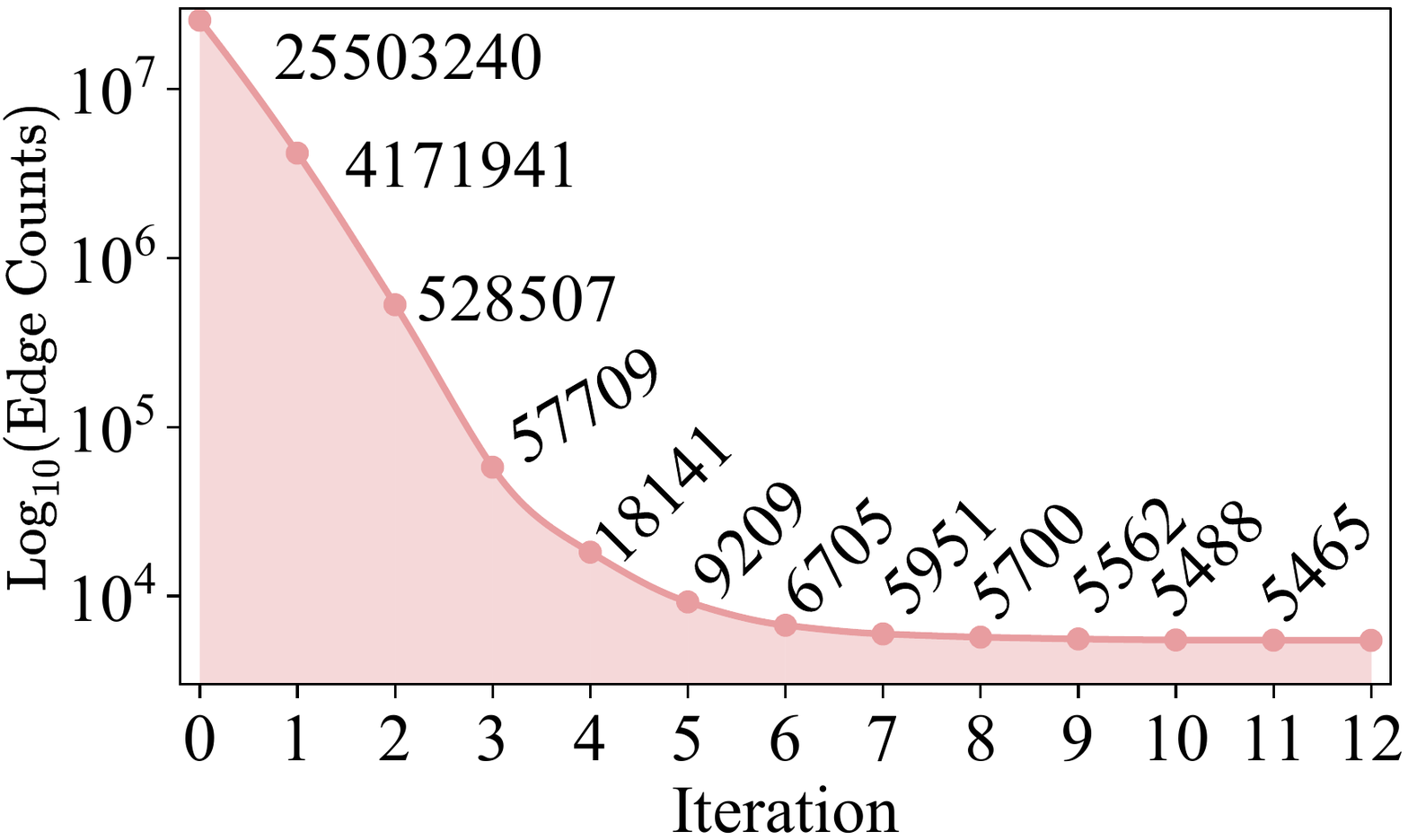}
        \vspace{-0.6cm}
    \end{subfigure}
   \begin{subfigure}{0.325\linewidth}
        \centering
        \includegraphics[width=\linewidth,trim={0 1cm 0 0}, clip]{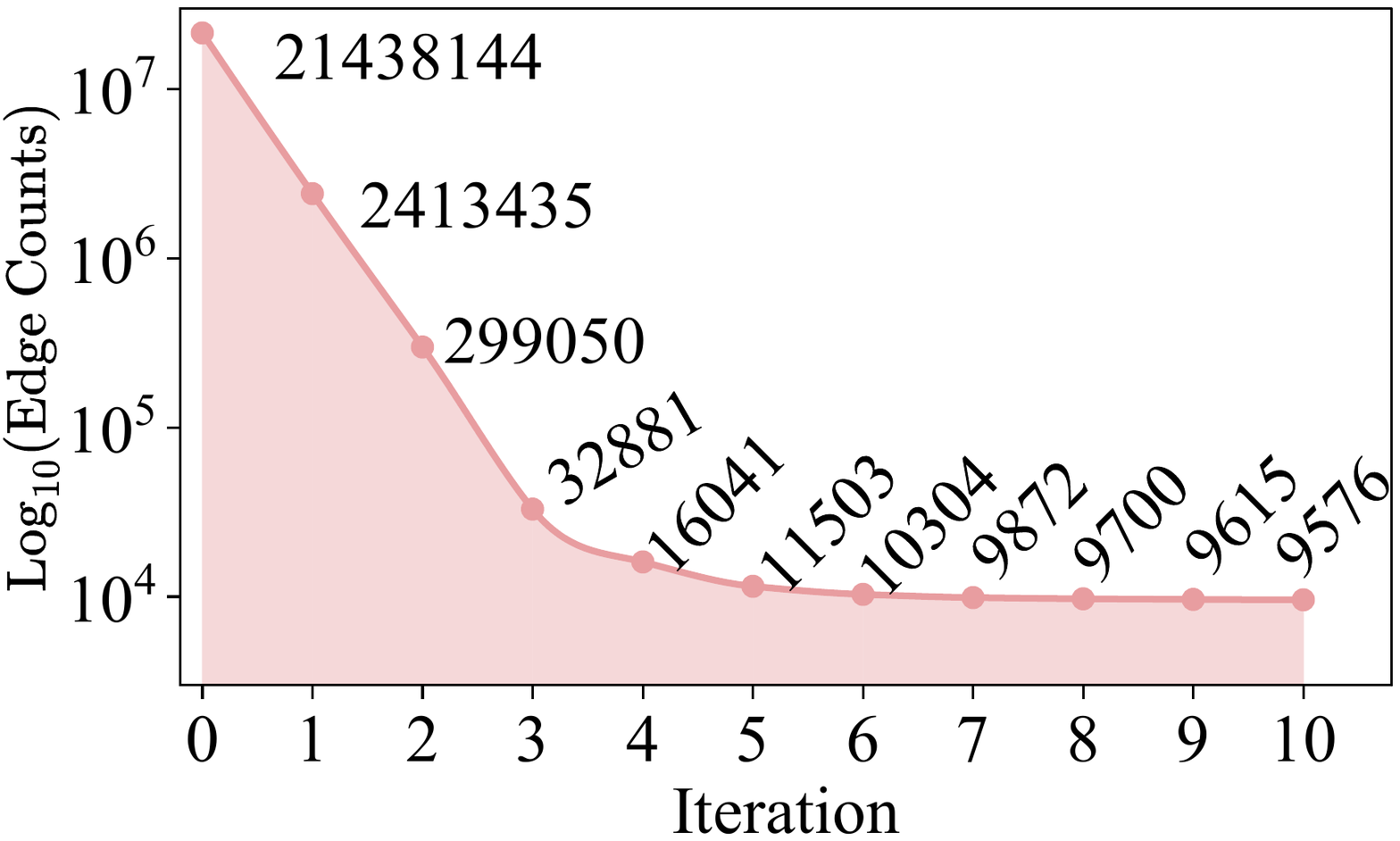}
        \vspace{-0.6cm}
    \end{subfigure}
    \begin{subfigure}{0.325\linewidth}
        \centering
        \includegraphics[width=\linewidth,trim={0 1cm 0 0}, clip]{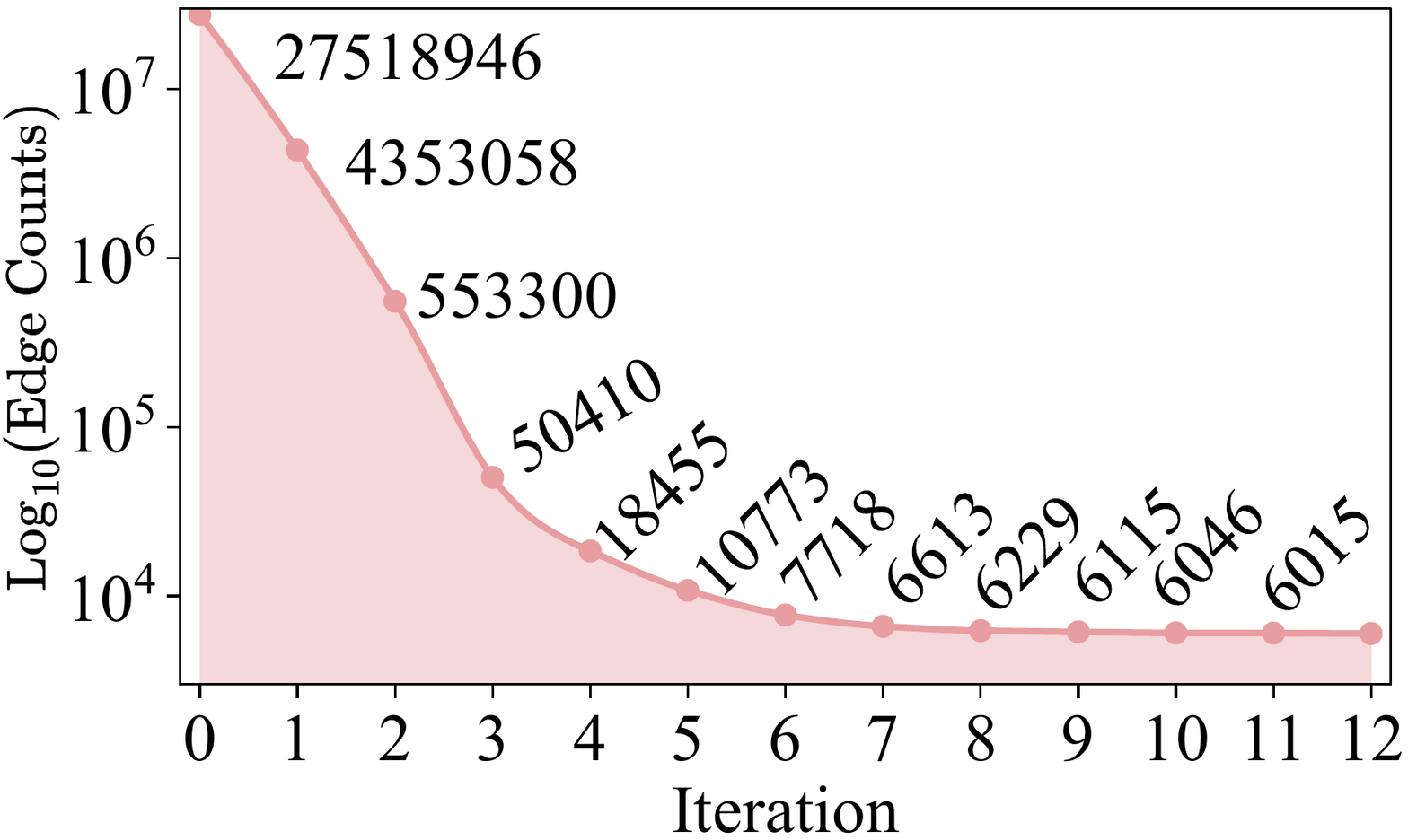}
        \vspace{-0.6cm}
    \end{subfigure}

    \begin{subfigure}{0.325\linewidth}
        \centering
        \includegraphics[width=\linewidth,trim={0 1cm 0 0}, clip]{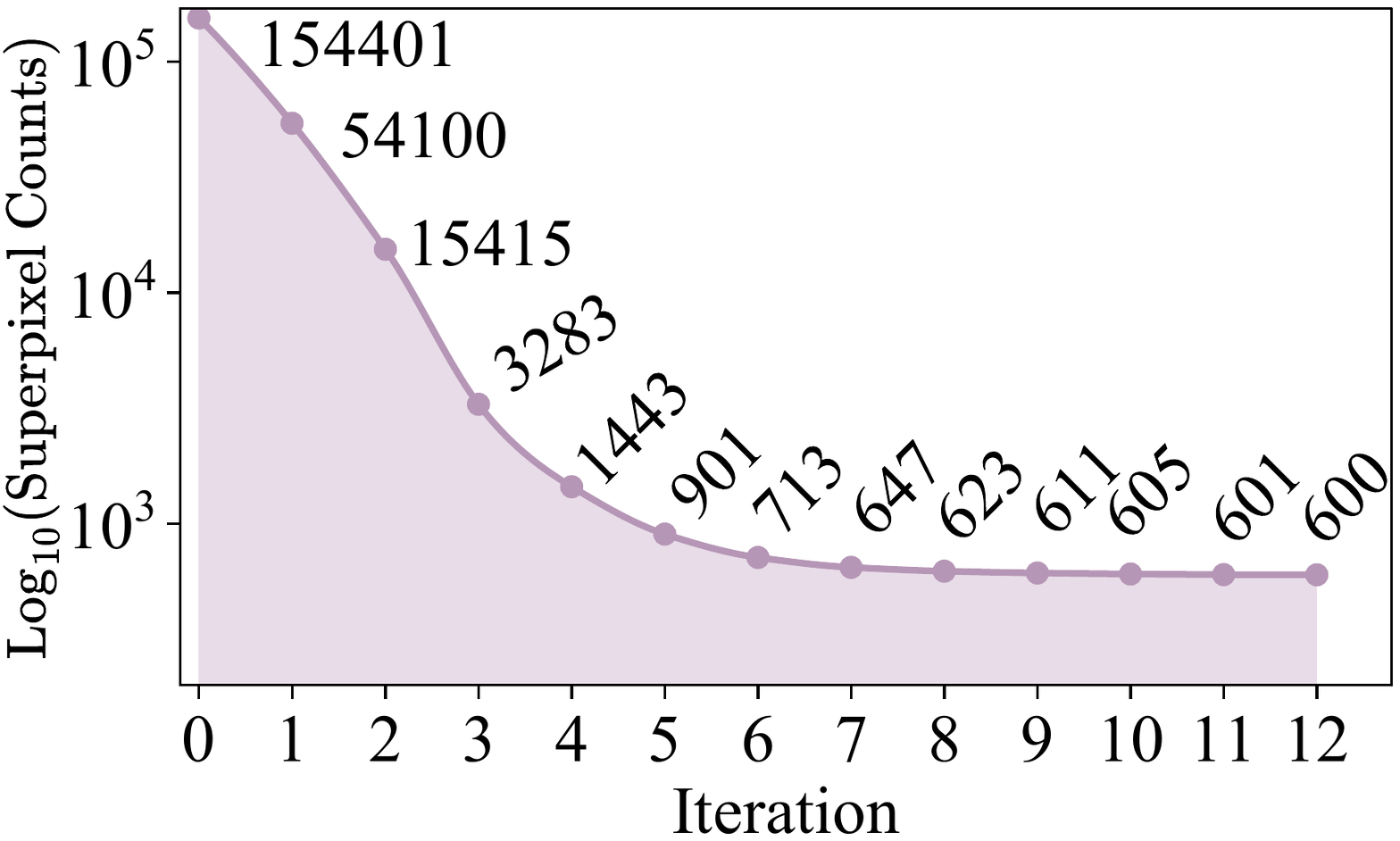}
        \vspace{-0.6cm}
        \caption{BSDS500}
    \end{subfigure}
    \begin{subfigure}{0.325\linewidth}
        \centering
        \includegraphics[width=\linewidth,trim={0 1cm 0 0}, clip]{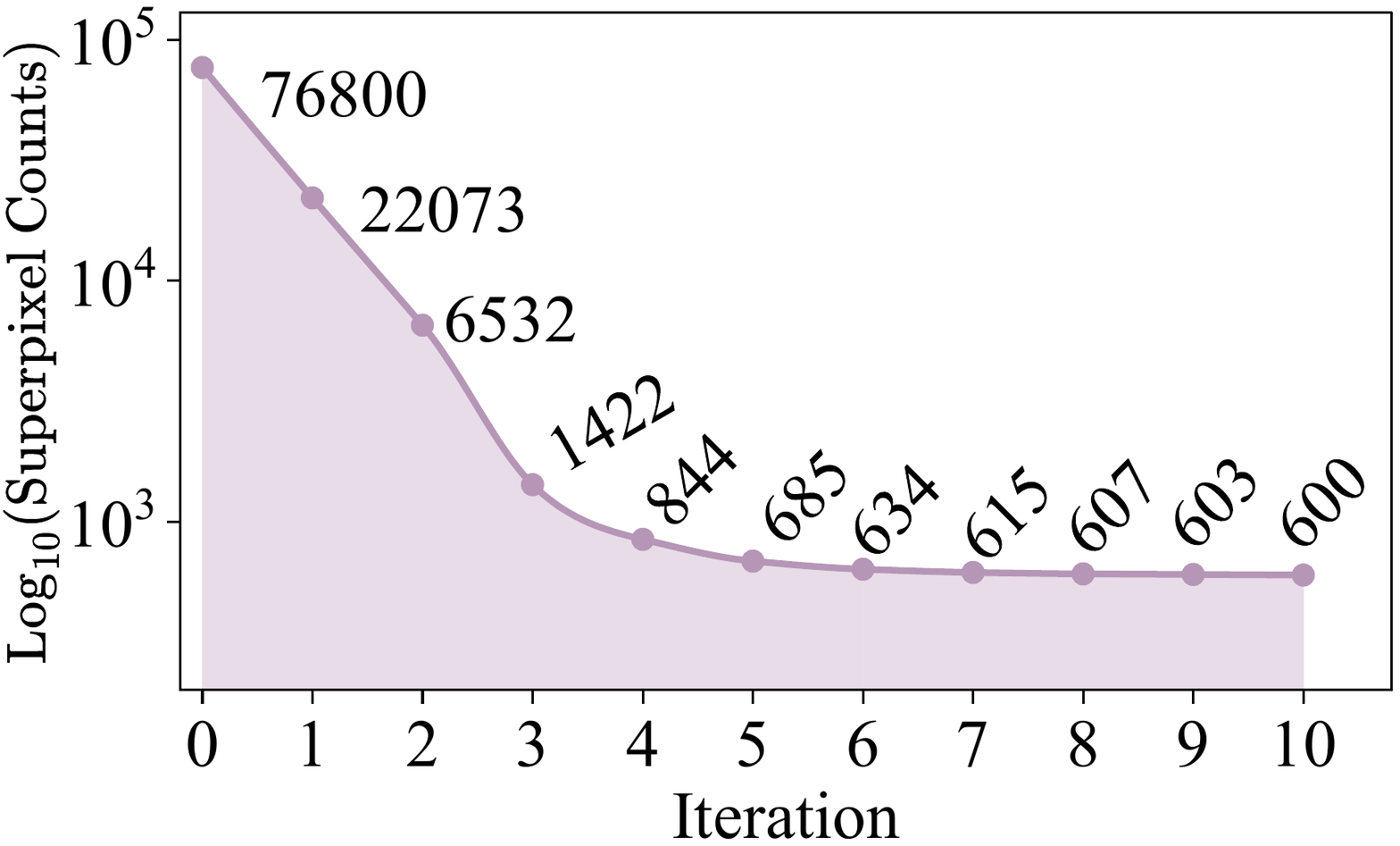}
        \vspace{-0.6cm}
        \caption{SBD}
    \end{subfigure}
    \begin{subfigure}{0.325\linewidth}
        \centering
        \includegraphics[width=\linewidth,trim={0 1cm 0 0}, clip]{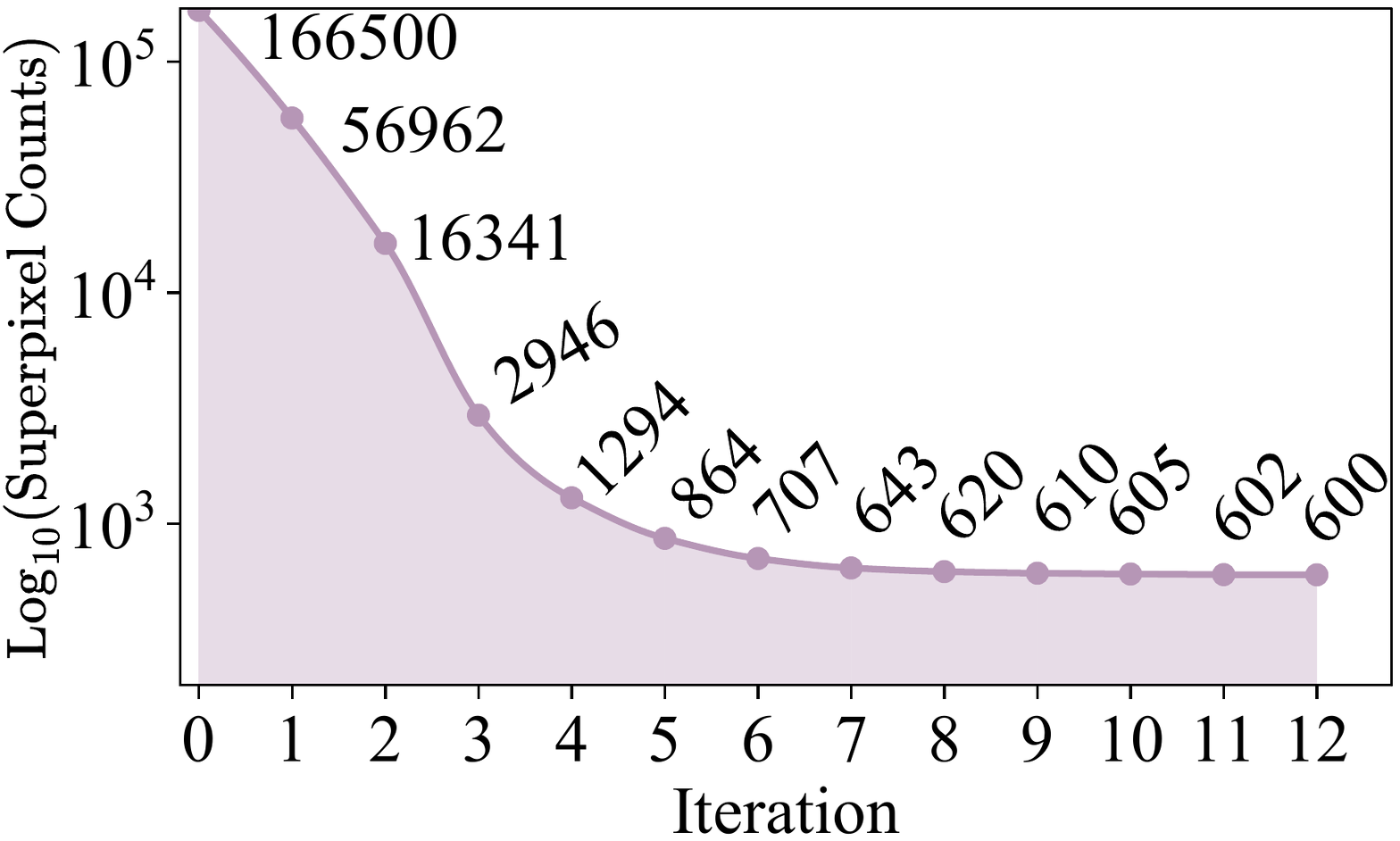}
        \vspace{-0.6cm}
        \caption{PASCAL-S}
    \end{subfigure}
    \caption{The changes in the  edges numbers and superpixel numbers between them at each iteration from images in BSDS500, SBD, and PASCAL-S datasets.}\label{fig:iteration}
\end{figure*}
\end{document}